\documentclass[runningheads]{llncs}

\usepackage{eccv}

\usepackage{eccvabbrv}

\usepackage{graphicx}
\usepackage{booktabs}

\usepackage{threeparttable}
\usepackage[table,dvipsnames]{xcolor} %
\usepackage{pifont}
\usepackage{caption}
\usepackage{mathtools, comment, multirow} 
\usepackage{ulem}
\usepackage{enumitem}

\definecolor{BestColor}{rgb}{1, 0.7, 0.7}
\definecolor{SecondColor}{rgb}{1, 0.85, 0.7}
\definecolor{ThirdColor}{rgb}{1, 1, 0.7}

\newcommand{\first}[1]{\cellcolor{BestColor}#1}
\newcommand{\second}[1]{\cellcolor{SecondColor}#1}
\newcommand{\third}[1]{\cellcolor{ThirdColor}#1}

\newcommand{\firsttext}[1]{\colorbox{BestColor}{#1}}
\newcommand{\secondtext}[1]{\colorbox{SecondColor}{#1}}
\newcommand{\thirdtext}[1]{\colorbox{ThirdColor}{#1}}

\usepackage[accsupp]{axessibility}  

\usepackage[colorlinks]{hyperref}

\usepackage{orcidlink}

\begin{document}

\title{4DGS360: 360° Gaussian Reconstruction of Dynamic Objects from a Single Video} 

\titlerunning{4DGS360}

\author{Jae Won Jang\inst{1}\orcidlink{0009-0008-4772-5144} \and
Yeonjin Chang\inst{1}\orcidlink{0009-0005-0735-2759} \and
Wonsik Shin\inst{1}\orcidlink{0009-0009-7458-9579} \and \\
Juhwan Cho\inst{1}\orcidlink{0009-0000-3337-0540} \and
Nojun Kwak\inst{1}\orcidlink{0000-0002-1792-0327} \thanks{Corresponding author.}}

\authorrunning{J.Jang et al.}

\institute{Seoul National University, Seoul, South Korea\\
\email{\{pert0407, yjean8315, wonsikshin, hj99cho, nojunk\}@snu.ac.kr}}

\maketitle

\begin{abstract}

We introduce 4DGS360, a diffusion-free framework for 360$^{\circ}$ dynamic object reconstruction from casual monocular video. Existing methods often fail to reconstruct consistent 360$^{\circ}$ geometry, as their heavy reliance on 2D-native priors causes initial points to overfit to visible surface in each training view. 4DGS360 addresses this challenge through an advanced 3D-native initialization that mitigates the geometric ambiguity of occluded regions. Our proposed 3D tracker, AnchorTAP3D, produces reinforced 3D point trajectories by leveraging confident 2D track points as anchors, suppressing drift and providing reliable initialization that preserves geometry in occluded regions. This initialization, combined with optimization, yields coherent 360$^{\circ}$ 4D reconstructions. We further present iPhone360, a new benchmark where test cameras are placed up to 135$^{\circ}$ apart from training views, enabling 360$^{\circ}$ evaluation that existing datasets cannot provide. Experiments show that 4DGS360 achieves state-of-the-art performance on the iPhone360, iPhone, and DAVIS datasets, both qualitatively and quantitatively. Project website at \url{https://jaewon040.github.io/4dgs360/}


\end{abstract}
\section{Introduction}
\label{sec:intro}

We present 4DGS360, a diffusion-free framework for 360$^{\circ}$ dynamic object reconstruction from casual monocular video. By leveraging 3D-native occlusion-aware initialization, our method ensures faithful 3D reconstruction even at extreme novel viewpoints.

Dynamic 3D reconstruction has long been a central area of interest in computer vision. 
Beyond static 3D Gaussian Splatting~\cite{kerbl20233d}, 4D (3D $+$ time) reconstruction is increasingly demanded in practical applications such as video content creation, spatial computing for VR/MR/AR, and 3D holographic media. In particular, reconstructing from a casual monocular video, a setting that reflects real-world capture conditions, has garnered significant attention~\cite{wu20254dfly, lei2025mosca}.

4D reconstruction under a monocular setting is a highly ill-posed problem~\cite{gao2022dycheck, lee2023fast, liu2023robust}, as only a single viewpoint is available per frame, with no multi-view stereo cues available.
Recent works~\cite{liang2025himor, wang2025shape} have addressed this challenge by leveraging pretrained 2D point tracking models~\cite{Doersch_2024_bootstap}, which establish cross-frame correspondences in dynamic videos. 

\begin{figure}[t]
  \centering
  \includegraphics[width=\textwidth]{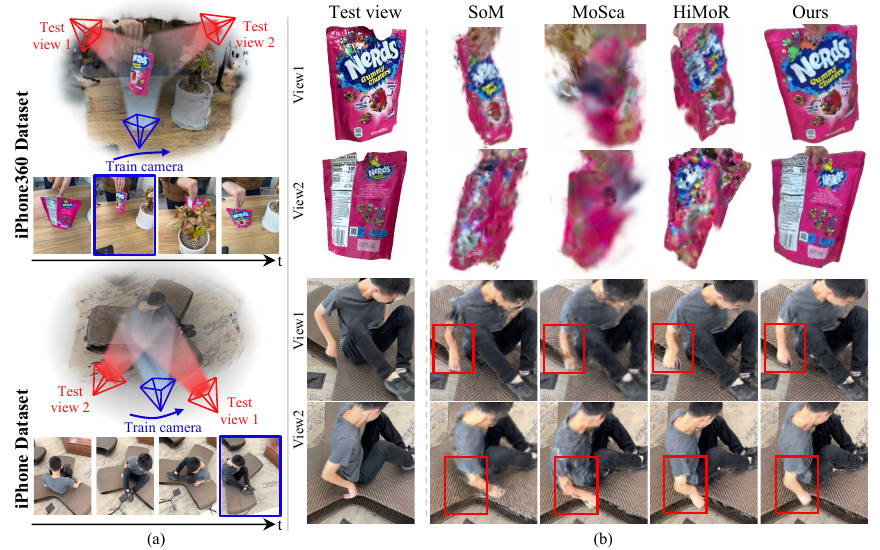}
  \caption{
  Our 4DGS360 model reconstructs 360° dynamic object geometry from a monocular video and produces higher-fidelity renderings than prior methods under both ordinary and extreme novel-view conditions. We also introduce the iPhone360 dataset, which captures dynamic objects in a monocular setting and includes test cameras far outside the training trajectory for 360° evaluation.
  } 
  \label{fig:teaser}
\end{figure}

However, as shown in \cref{fig:teaser}, existing methods still fail to reconstruct regions observed at extreme novel viewpoints (e.g., $>90^\circ$ from the current view), even when those regions are visible in other frames of the input video.
We argue that this failure stems from a heavy reliance on 2D-native priors~\cite{Doersch_2024_bootstap, doersch2023tapir} during initialization. Existing methods~\cite{lei2025mosca, liang2025himor} initialize 3D dynamic trajectories using 2D tracking results, which offer more reliable performance than 3D tracking models~\cite{tapip3d, xiao2025spatialtrackerv2}. These 2D tracks must be lifted into 3D to form initial gaussians. However, depth maps used in lifting only provide depth for surfaces visible in the current frame. As a result, the 3D positions of occluded track points remain ambiguous, causing the initialized geometry to overfit to the visible surfaces at each timestep. Since this incomplete geometry is not recovered during optimization, reconstruction of occluded regions fails, even when they are visible in other frames.

We propose 4DGS360 to address the longstanding challenge of complete 360° dynamic object reconstruction by introducing AnchorTAP3D, a novel 3D native tracker that overcomes the limitations of both 2D and 3D tracking models effectively. AnchorTAP3D leverages confident 2D track points as anchors for 3D tracking, improving long-term reliability and resolving the depth ambiguity of occluded regions without additional training. AnchorTAP3D provides a reliable initialization that unlocks the full potential of the optimization strategy used in prior works. In particular, the As-Rigid-As-Possible (ARAP) regularization, which previously failed to operate correctly on corrupted initialized geometry in occluded regions, can now function as intended, enabling significantly improved 360° reconstructions.
As shown in \cref{fig:teaser}(b), our method establishes more stable and coherent 4D reconstructions of dynamic objects, even under large view extrapolation.

However, existing monocular datasets~\cite{park2021hypernerf, park2021nerfies, gao2022dycheck} for dynamic scene reconstruction are limited to evaluate reconstruction quality at truly novel viewpoints, as they lack sufficient view disparity between train and test cameras. Even the iPhone dataset~\cite{gao2022dycheck}, which provides the largest such disparity, remains insufficient to properly evaluate 360° reconstruction.
To address this, we introduce iPhone360, a new dataset designed to evaluate 360° reconstruction of dynamic objects under realistic conditions with large disparity between train and test camera views. Our iPhone360 dataset includes training videos of real-world dynamic objects captured with casual handheld movements, exhibiting diverse motion ranging from object manipulation to human motion. 
\cref{fig:teaser}(a) demonstrates how our iPhone360 dataset provides significantly larger view disparity compared to the iPhone dataset, enabling 360$^{\circ}$ evaluation.

\noindent Overall, we provide the following contributions:
\begin{itemize}
    \item We propose \textbf{4DGS360}, which employs a novel 3D-native initialization method based on \textit{AnchorTAP3D}, enabling consistent 360° dynamic reconstruction from monocular video.
    \item We present \textbf{iPhone360 dataset}, a new dataset that captures real-world dynamic objects with various scenarios. Test cameras are placed up to 135° apart from training views enabling evaluation of models under extreme novel-view conditions.
    \item Our approach achieves state-of-the-art performance both qualitatively and quantitatively under ordinary and extreme novel-view synthesis conditions.
\end{itemize}


\section{Related Work}
\label{sec:related work}

\subsection{Dynamic Novel-View Synthesis}
Novel-view synthesis has rapidly advanced with the advent of NeRF~\cite{mildenhall2021nerf} and 3D Gaussian Splatting (3DGS)~\cite{kerbl20233d}.  
NeRF represents scenes implicitly using multi-layer perceptrons (MLPs), enabling view-consistent rendering across various reconstruction tasks~\cite{barron2021mip, barron2022mip,seo2023flipnerf,wang2022clip,verbin2024ref,yu2021pixelnerf,lee2025divcon,barron2023zip,duckworth2024smerf}.  
In contrast, 3DGS models scenes explicitly with 3D Gaussian primitives, enabling real-time rasterization-based rendering through its efficient explicit formulation. Owing to these advantages, the 3DGS framework has been widely extended to diverse applications~\cite{chu2024dreamscene4d, lu2024scaffold, mallick2024taming, Shen2024SuperGaussian, Yu2024MipSplatting, Chen_deblurgs2024, zhao2024badgaussians, chen2024mvsplat, liu2025mvsgaussian, paliwal2024coherentgs, charatan23pixelsplat, hu2024evagaussian3dgaussianbasedrealtime}.

Extending these ideas to dynamic scenes, models reconstruct objects and environments that change over time.  
Recent works~\cite{liang2025himor,wang2025shape} model temporal deformation by estimating the trajectories of canonical-space Gaussians.  
Some methods~\cite{wu20244dgs,yang2024deformable} encode motion implicitly via MLPs, while others represent motion explicitly by assigning trajectories to individual Gaussians~\cite{stearns2024marbles,luiten2024dynamic, wu20254dfly} or by combining learned bases~\cite{wang2025shape} and hierarchical motion fields~\cite{liang2025himor}.

Recent studies further explore the monocular setting, where a single camera captures dynamic scenes under realistic conditions.  
However, monocular reconstruction remains highly ill-posed and typically relies on pretrained models~\cite{yang2023track, karaev23cotracker, karaev24cotracker3, depth_anything_v2, depthanything, Wang_2024_CVPR, 10.5555/3540261.3541527, Kirillov_2023_ICCV, Harley2022ParticleVR} or 2D tracking cues~\cite{doersch2023tapir, Doersch_2024_bootstap,wang2025gflow,liu2025modgs} to recover motion on the image plane.
While effective for visible regions, these approaches fail to reconstruct occluded geometry, leading to overfitting to the training views.  
Other works~\cite{wu2025difix3d+, sam3dteam2025sam3d3dfyimages, kim20244d} employ large diffusion models~\cite{rombach2022high} to synthesize unseen view. However, these methods require substantial computational cost for training generative model, and often fail to leverage information across video frames. Furthermore, as demonstrated in \cref{fig:difix}, they show limited performance gains even when combined with existing state-of-the-art 4D novel view synthesis methods at extreme novel viewpoints.

Our method addresses this limitation by introducing occluded geometry-aware tracking at initialization, enabling consistent reconstruction beyond observed views and serving as a stronger baseline even when combined with diffusion-based approaches.

\subsection{Monocular setting datasets}
A number of benchmarks have been proposed for dynamic scene reconstruction under the monocular setting.  
D-NeRF~\cite{pumarola2021dnerf} uses synthetic scenes to evaluate temporal deformation modeling, while HyperNeRF~\cite{park2021hypernerf}, Nerfies~\cite{park2021nerfies}, and the iPhone dataset~\cite{gao2022dycheck} capture real-world dynamic scenes.  
These datasets mainly evaluate temporal interpolation or short-range novel view synthesis, where test cameras remain close to the train camera.  
In particular, Hypernerf and Nerfies datasets are not fully aligned with real-world capture conditions. They alternate between two cameras per frame to construct training sets. The iPhone dataset maintains a more faithful monocular setting, but the gaps between train and test cameras are limited. 
Therefore, we release a new dataset, \textbf{iPhone360}, for comprehensive 360° evaluation with extreme novel view test cameras under realistic monocular capture conditions.


\subsection{2D and 3D Point Tracking}

\paragraph{2D Tracking.}
Recent point tracking methods adopt deep learning approaches. Among them, TAPIR~\cite{doersch2023tapir} improves accuracy through global matching and refinement, while transformer-based models such as CoTracker~\cite{karaev24cotracker3} iteratively infer point position and visibility, achieving robustness under occlusion. Self-supervised approaches like BootsTAP~\cite{Doersch_2024_bootstap} further enhances performance on unlabeled real-world data using pseudo ground truth from pretrained trackers.

 \paragraph{3D Tracking.}
Among 3D tracking models~\cite{karaev23cotracker, karaev24cotracker3}, TAPIP3D~\cite{tapip3d} achieves notable tracking results by employing a transformer-based architecture that performs tracking directly in XYZ space, leveraging unprojected image features to construct a spatio-temporal 3D feature cloud. It models temporal correspondence through 3D neighborhood-to-neighborhood attention, where local geometric features near query and target points are jointly attended to infer motion trajectories. This design enables TAPIP3D to capture smooth and continuous motion across frames and handle moderate occlusions without explicit depth supervision.
While 3D tracking provides more geometric understanding than 2D tracking models, it is sensitive to errors in depth maps and camera calibration, and tracking failures tend to accumulate over time, reducing long-term stability. 
On the other hand, 2D tracking is generally more resilient to noise in real-world settings but provides limited spatial understanding.
To address this, we propose \textbf{AnchorTAP3D}, which leverages high confidence 2D track points as anchors for 3D tracking, enabling consistent geometry-aware initialization for our model.

\section{Method}
\label{sec:method}
\newcommand{\framework}{MyMethod}
\begin{figure*}[t]
    \centering
    \includegraphics[width=\linewidth]{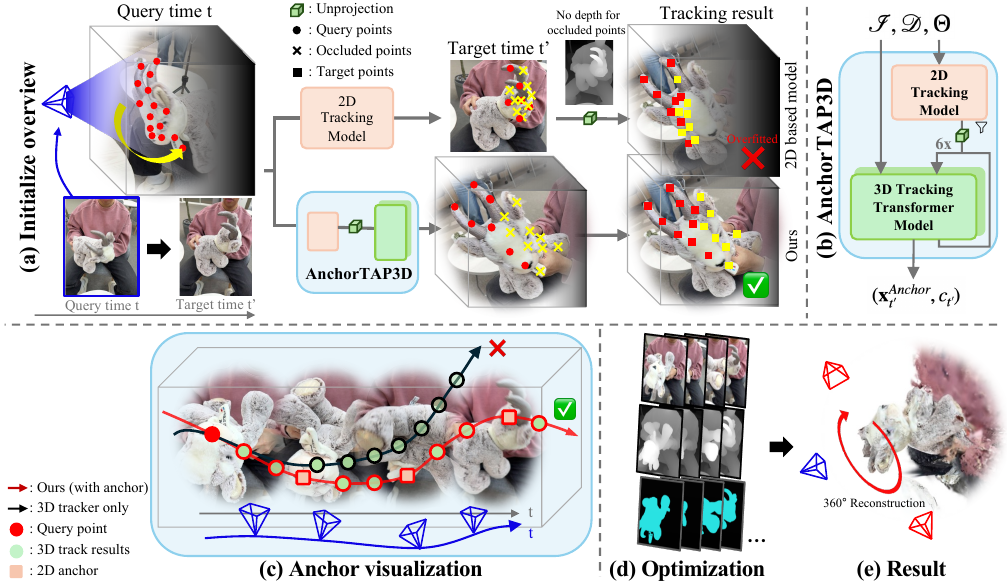}
    \caption{
    \textbf{Overview of 4DGS360} (a) illustrates the initialization stage using AnchorTAP3D, where reliable geometry can be obtained even when tracked points become occluded. This contrasts with 2D-based models, which cannot estimate depth for occluded points. (b) presents the overall architecture of AnchorTAP3D, and (c) highlights the difference between tracking with and without anchors, showing that anchors effectively prevent error drift. (d) depicts the training process that follows initialization, and (e) shows that our model finally enables full $360^{\circ}$ reconstruction. The blue pyramids represent training cameras, while the red pyramids indicate test cameras.
    }
    \label{fig:method}
\end{figure*}

Our goal is to reconstruct a 360\textdegree~4D representation over time, 
given a sequence of $T$ training frames 
$\mathcal{I}=\{I_t\}_{t=1}^{T}$, depth maps $\mathcal{D}=\{D_t\}_{t=1}^{T}$, and camera parameters $\Theta=\{\theta_t\}_{t=1}^{T}$
obtained either from dataset sensors or pretrained estimators.
Our method represents dynamic objects using a set of static 3D Gaussians defined in a canonical space, which are deformed over time according to a hierarchical motion structure. Sec.~\ref{subsec:representation} describes the preliminaries and details our 4D scene representation.
Sec.~\ref{subsec:initialization} introduces our occluded geometry–aware initialization using the proposed tracking model, AnchorTAP3D. 
Finally, Sec.~\ref{subsec:optimization} presents the optimization strategy that captures dynamic motion while enforcing local rigidity and visual consistency. \cref{fig:method} illustrates the overview of our method.

\subsection{Preliminary: Dynamic Gaussian Splatting}
\label{subsec:representation}

A 3D scene is represented by a set of anisotropic Gaussian primitives 
$G(\boldsymbol{\mu}, \boldsymbol{\Sigma}, \alpha, \mathbf{c})$, 
where each Gaussian is parameterized by its mean 
$\boldsymbol{\mu} \in \mathbb{R}^3$, 
covariance 
$\boldsymbol{\Sigma} \in \mathbb{R}^{3\times3}$, 
opacity $\alpha \in \mathbb{R}^+$, 
and view-dependent color coefficients 
$\mathbf{c} \in \mathbb{R}^{3(l+1)^2}$ 
from spherical harmonics (SH). 
This explicit and differentiable formulation enables photorealistic scene representation and optimization in 3D space.

To handle dynamics, each static Gaussian in a canonical frame is deformed over time through 
$\mathbf{T}_t = [\mathbf{R}_t | \mathbf{t}_t] \in \mathrm{SE}(3)$, 
resulting in the time-dependent primitive
\begin{equation}
G_t = G(\mathbf{R}_t \boldsymbol{\mu}_0 + \mathbf{t}_t,\, \mathbf{R}_t \boldsymbol{\Sigma}_0\mathbf{R}_t^{\top},\, \alpha,\, \mathbf{c}),
\label{eq:dynamic_gaussian}
\end{equation}
where $(\boldsymbol{\mu}_0, \boldsymbol{\Sigma}_0)$ denote the mean and covariance at canonical space. 

We model $\mathbf{T}_t$ following the HiMoR~\cite{liang2025himor}, which encodes the deformation field as a tree of nodes.
The deformation of a canonical Gaussian is guided by its nearby leaf nodes. 
Specifically, the transformation $\mathcal{T} = \{\, \mathbf{T}_t \,\}_{t=1}^{T}$ of a Gaussian $G$
is obtained by interpolating the motions of its $K$ nearest leaf nodes $\mathcal{V}$:
\begin{equation}
\mathbf{T}_t = 
\sum_{k \in \mathcal{N}(G,\mathcal{V})} 
w_k \, \mathcal{M}_{k},
\label{eq:interp}
\end{equation}
where $\mathcal{M} = \{\, \mathbf{M}_t \,\}_{t=1}^{T}$ for $\mathbf{M}_t \in \mathrm{SE}(3)$ denotes the motion of nodes, 
and $w_k$ are Gaussian-specific interpolation weights.

Each node’s motion $\mathbf{M}_t$ represents its transformation from the canonical frame to frame $t$, 
and is defined hierarchically with respect to its parent. 
To efficiently encode these motions, HiMoR represents each node’s transformation 
as a weighted combination of its parent’s shared motion bases:
\begin{equation}
\mathbf{M}_t = \sum_{m=1}^{M} v_m\, \mathcal{B}_{m,t},
\label{eq:basis}
\end{equation}
where $\{\mathcal{B}_{m,t}\}$ denote the motion bases shared among sibling nodes, 
and $v_m$ are node-specific coefficients. 
This hierarchical formulation allows higher-level nodes to capture global, smooth motion patterns, 
while deeper nodes refine fine-grained local deformation across time and space.

Given camera parameters $\boldsymbol{\theta}_t$, 
the deformed canonical Gaussian $G_t$ at time $t$ is projected onto the image plane via
\begin{equation}\label{eq:proj}
\begin{aligned}
\boldsymbol{\mu}'_t &= \Pi(\mathbf{R}_t \boldsymbol{\mu}_0 + \mathbf{t}_t; \boldsymbol{\theta}_t), \quad
\boldsymbol{\Sigma}'_t = \Pi(\mathbf{R}_t \boldsymbol{\Sigma}_0\mathbf{R}_t^{\top}; \boldsymbol{\theta}_t),
\end{aligned}
\end{equation}
where $\Pi$ denotes the camera projection. 
After projection, each 3D Gaussian $G_t$ becomes a 2D Gaussian on the image plane, 
parameterized by the mean $\boldsymbol{\mu}'_t \in \mathbb{R}^2$ and covariance 
$\boldsymbol{\Sigma}'_t \in \mathbb{R}^{2\times2}$.

The rendered color at pixel $\mathbf{p} \in \mathbb{R}^2$ is then computed 
by aggregating these 2D Gaussians using depth-sorted alpha blending:
{
\begin{equation}\label{eq:render}
\begin{split}
C_t(\mathbf{p}) &=
\sum_{i=1}^{N}
\mathbf{c}_i\, \sigma_{i,t}(\mathbf{p})
\prod_{j=1}^{i-1}\!\left(1-\sigma_{j,t}(\mathbf{p})\right),\\
\sigma_{i,t}(\mathbf{p}) &=
\alpha_i \exp\!\left(
-\tfrac{1}{2}\,
(\mathbf{p}-\boldsymbol{\mu}'_{i,t})^\top
(\boldsymbol{\Sigma}'_{i,t})^{-1}
(\mathbf{p}-\boldsymbol{\mu}'_{i,t})
\right),
\end{split}
\end{equation}
}
where $\sigma_{i,t}(\mathbf{p})$ denotes the 2D Gaussian opacity at pixel $\mathbf{p}$, 
and $\mathbf{c}_i$ is its view-dependent color. 
This dynamic Gaussian representation enables a differentiable rasterization framework.

\subsection{Initialization Method}
\label{subsec:initialization}
\paragraph{AnchorTAP3D.} 
We introduce \textbf{AnchorTAP3D} (\textbf{Anchor}-guided \textbf{T}racking \textbf{A}ny \textbf{P}oint in \textbf{3D}), 
an advanced 3D tracking model designed to enhance dynamic $360^\circ$ reconstruction under extreme novel views 
without additional training. 

\begin{figure}[t]
    \centering
    \includegraphics[width=\linewidth]{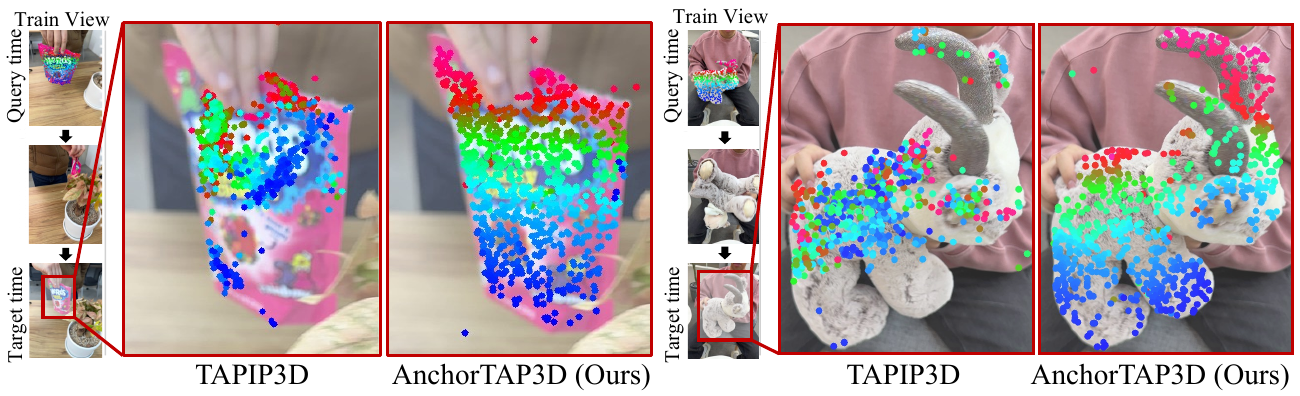}
    \caption{
    \textbf{Tracking Comparison}. We compare the naive 3D tracking model TAPIP3D~\cite{tapip3d} with our anchor-based AnchorTAP3D (Ours). At the target time after undergoing rotation and occlusion, AnchorTAP3D produces noticeably more stable tracking results. This indicates that anchor-guided tracking is advantageous under challenging conditions such as occlusion.
    }
    \label{fig:anchortap3d}
\end{figure}

Given a sequence of frames $\mathcal{I}$, depth maps $\mathcal{D}$, and camera parameters $\Theta$,
AnchorTAP3D estimates the 3D location $\mathbf{x}^{Anchor}_{t'} \in \mathbb{R}^3$ of a 2D anchor point $\mathbf{p}_t \in \mathbb{R}^2$ observed at time $t$, after it has moved to time $t'$, along with its binary visibility $v_{t'}$:
\begin{equation}
(\mathbf{x}^{Anchor}_{t'}, v_{t'}) =
\mathcal{P}^{Anchor3D}_{t\rightarrow t'}(\mathbf{p}_t, \mathcal{I}, \mathcal{D}, \Theta),
\label{eq:anchor3d_overview}
\end{equation}
where $\mathcal{P}^{Anchor3D}$ denotes our unified anchor-guided model which integrates the strengths of a 2D tracking model and a 3D tracking model into a unified framework. The overall formulation can be decomposed into several components, which we describe below.

The 2D tracker, $\mathcal{P}^{2D}$, 
takes a query point 
$\mathbf{p}_{t} = (x_t, y_t)$ at time $t$ and predicts its correspondence in the target frame $t'$ as:
\begin{equation}
(\mathbf{p}_{t'}, c_{t'}) = \mathcal{P}^{2D}_{t\rightarrow t'}(\mathbf{p}_{t}, \mathcal{I}),
\end{equation}
where $c_{t'}$ denotes the tracking confidence estimated by the 2D tracker.

The predicted 2D correspondence is then lifted to 3D space through inverse camera projection using the depth map and camera parameters at frame $t'$:
\begin{equation}
\mathbf{x}^{2D}_{ t'} = \Pi^{-1}_{t'}(\mathbf{p}_{t'}, D_{t'}, \boldsymbol{\theta}_{t'}).
\end{equation}
This process generates candidate 3D points corresponding to each tracked 2D point. However, points with low 2D confidence, mostly occluded, cannot obtain reliable 3D positions from the depth map of the target frame.

To exclude unreliable correspondences, we define a binary mask based on the 2D confidence score:
\begin{equation}
\text{Mask}_{t'} =
\begin{cases}
1, & c_{t'} > \tau,\\
0, & \text{otherwise}.
\end{cases}
\label{eq:mask}
\end{equation}
Only points satisfying $c_{t'} > \tau$ are regarded as trustworthy and are used to form anchor points for subsequent 3D inference.

Within a sliding temporal window $w(t,t')$ of fixed length $L$, 
the transformer-based 3D tracker jointly processes all frames in the window 
(e.g., $L\!=\!16$, with $8$-frame overlap between successive windows). 
During each inference step, we collect reliable 3D anchor points 
obtained from high-confidence 2D tracks as:
\begin{equation}
\mathcal{X}^{A}_{w(t,t')} \;=\;
\big\{\, \mathbf{x}^{2D}_{s'} \;\big|\;
s'\!\in\! w(t,t'),\; \text{Mask}_{s'}\!=\!1
\,\big\}.
\label{eq:anchorset}
\end{equation}
Here, $\mathcal{X}^{A}_{w(t,t')}$ denotes the set of 3D anchor points that condition the transformer during the current temporal inference window, 
providing geometry-consistent supervision across frames (See \cref{fig:method}(c)).

Finally, the anchor-guided 3D tracker predicts the target 3D point and its binary visibility score by conditioning on the anchor set:
\begin{equation}
(\mathbf{x}^{Anchor}_{t'}, v_{t'}) =
\mathcal{P}^{3D}_{t\rightarrow t'}(\mathbf{p}_t, \mathcal{I}, \mathcal{D}, \Theta, \mathcal{X}^{A}_{w(t,t')}).
\end{equation}

Unlike conventional 3D trackers that propagate a single query over time, AnchorTAP3D leverages multiple anchors as spatial-temporal constraints, 
suppressing error drift and enhancing geometric stability. 
The anchors are relatively unaffected by depth or calibration noise during tracking,
providing stable initialization for our model. \cref{fig:anchortap3d} illustrates that our model maintains long-term tracking more effectively than a naive 3D tracking approach. Furthermore, \cref{fig:ablation} demonstrates that in the 4D reconstruction task, our method initialized with AnchorTAP3D better reconstructs the geometry of occluded regions compared to baselines using naive 2D or 3D tracking.

\paragraph{Initialization details.}

We initialize a set of dynamic Gaussians from the 3D tracks obtained at $T$ query times. 
Specifically, $N$ trajectories are randomly sampled to define Gaussian primitives whose positions and orientations vary over time. 

The frame with the most visible Gaussians is set as the canonical frame. 
We then cluster the trajectories according to their temporal velocities using $k$-means into $B$ groups. 
Within each cluster, frame-to-frame rigid transformations are estimated via Procrustes alignment~\cite{Schonemann_1966}, and the resulting temporal motions are stored as initial motion bases $\{\mathcal{B}_i\}_{i=1}^{B}$. 
Each Gaussian’s motion is further weighted by its spatial distance to the corresponding cluster center. 
Previous methods~\cite{wang2025shape, liang2025himor} discard invisible points during motion estimation since 2D tracking combined with depth-based unprojection cannot infer 3D positions for occluded points.
Our approach leverages AnchorTAP3D to infer plausible 3D positions even for occluded regions.
This capability enables motion bases that fully capture global object dynamics and establishes a foundation for complete 360° reconstruction.

For node initialization, we perform weighted random sampling of $n_1$ Gaussians from the canonical frame. 
Sampling weights are determined by both motion magnitude and spatial density, 
encouraging nodes to capture highly dynamic regions while maintaining even spatial coverage. 
The selected Gaussians initialize the first-level nodes with their positions and motion coefficients,
and the higher-level nodes are initialized in the same manner with respect to their parent nodes.

\subsection{Optimization Method}
\label{subsec:optimization}
To optimize the initialized nodes and Gaussians, we employ rigidity regularization to preserve geometric structure and render-based regularization that compares rendered outputs with ground truth.

\paragraph{Rigidity regularization.}
To maintain long-term geometric consistency and locally rigid motion across non-adjacent frames, 
we employ a generalized As-Rigid-As-Possible (ARAP)~\cite{arap} regularization.
For each node pair $(i, j)$ belonging to the same locally rigid cluster, 
we encourage both the pairwise distances and the relative local transformations 
to remain consistent between arbitrary frames $t$ and $t'$:

{\small
\begin{equation}
\mathcal{L}_{\text{arap}} = w_1 \sum_{(i,j) \in \mathcal{N}} \Big| \|\mathbf{x}_i^t - \mathbf{x}_j^t\|_2 - \|\mathbf{x}_i^{t'} - \mathbf{x}_j^{t'}\|_2 \Big| + w_2 \sum_{(i,j) \in \mathcal{N}} \Big\| \mathbf{T}_j^{t^{-1}}(\mathbf{x}_i^t) - \mathbf{T}_j^{t'^{-1}}(\mathbf{x}_i^{t'}) \Big\|_2,
\label{eq:arap}
\end{equation}
}
where $\mathbf{x}_i^t$ denotes the position of node $i$ at frame $t$, 
and $\mathbf{T}_j^t \in SE(3)$ represents the local transformation of node $j$.
This rigidity constraint allows temporally coherent propagation of structural information, 
enabling 360° geometry even under large motion or occlusion.

\paragraph{Render-based regularization.}
For render-based regularization, we conduct RGB loss $\mathcal{L}_{\text{rgb}}$, which includes D-SSIM~\cite{article_ssim} loss and LPIPS~\cite{zhang2018unreasonable} loss. We regularize the spatial compactness of the reconstructed object via a mask regularization loss $\mathcal{L}_{\text{mask}}$,  
enforce geometric alignment through a depth consistency loss $\mathcal{L}_{\text{depth}}$,  
and improve temporal correspondence using a 2D tracking loss $\mathcal{L}_{\text{2dtrack}}$.  
Together with the rigidity loss $\mathcal{L}_{\text{arap}}$ in Eq.~(\ref{eq:arap}), these objectives contribute to a temporally coherent and geometrically stable reconstruction:
{
\small
\begin{equation}
\begin{aligned}
\mathcal{L}_{\text{total}} =
&\;\lambda_{\text{rgb}} \mathcal{L}_{\text{rgb}}
+ \lambda_{\text{mask}} \mathcal{L}_{\text{mask}}
+ \lambda_{\text{depth}} \mathcal{L}_{\text{depth}} 
+ \lambda_{\text{2dtrack}} \mathcal{L}_{\text{2dtrack}}
+ \lambda_{\text{arap}} \mathcal{L}_{\text{arap}}.
\end{aligned}
\label{eq:total_loss}
\end{equation}
}
Further details are included in the supplementary.

\begin{figure*}
    \vspace{-1.em}
    \centering
    \includegraphics[width=\linewidth]{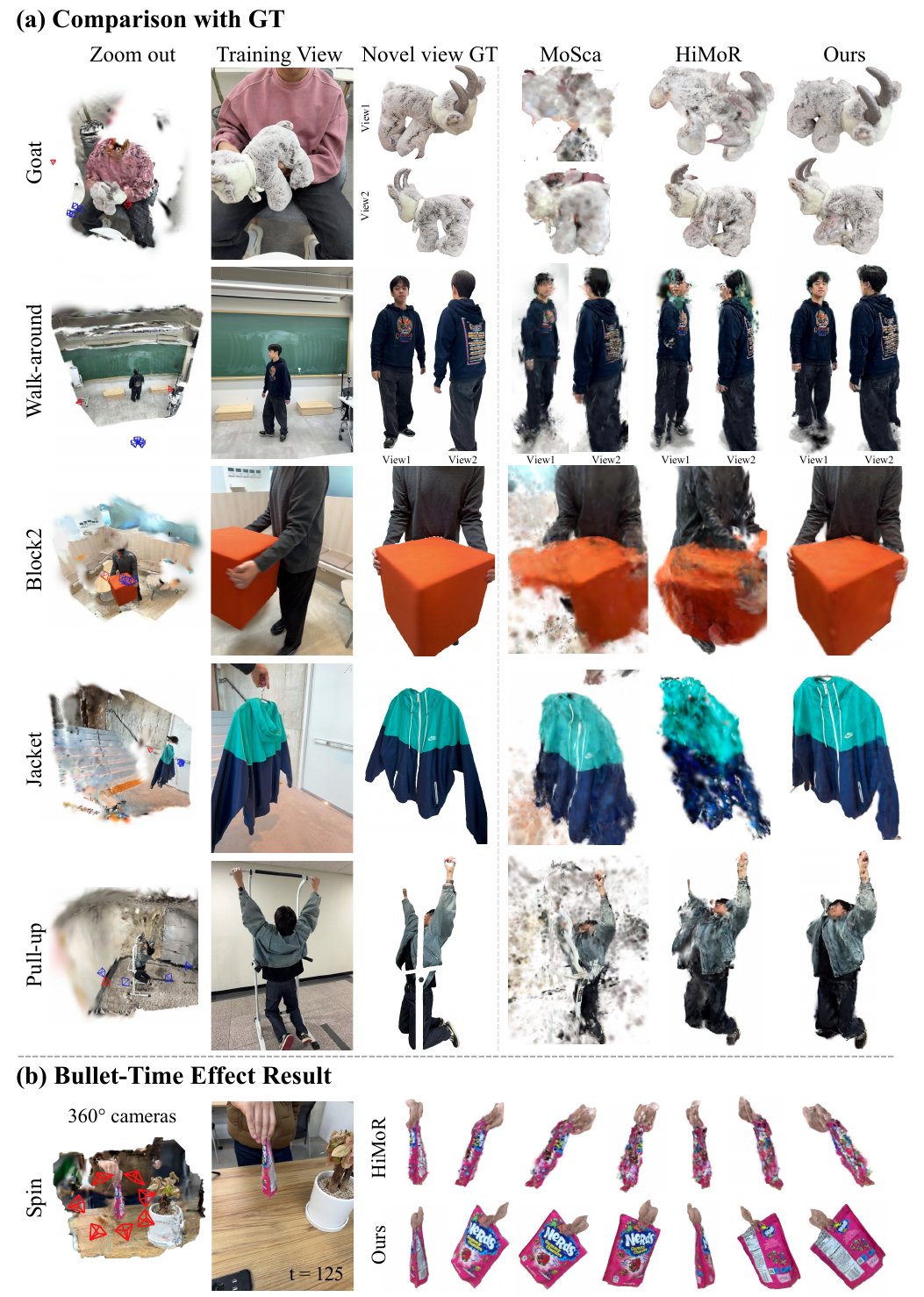}
    \vspace{-2.em}
    \caption{
    \textbf{iPhone360 Qualitative Comparison}. In (a), Novel-view GTs are masked and model outputs are shown as renderings of dynamic Gaussians. Blue pyramids denote training cameras, and reds denote test camera locations in the \textit{Zoom-out}. (b) shows bullet-time rendering results across 360° at a fixed time instant.
    }
    \label{fig:iphone360_qualitative}
    \vspace{-0.5em}
\end{figure*}

\section{Experiments}
\label{sec:experiments}

\newcommand{\cmark}{\textcolor{green}{\ding{51}}} 
\newcommand{\xmark}{\textcolor{red}{\ding{55}}}  

\subsection{Datasets and Evaluation}
\paragraph{Evaluation metrics.}
For evaluation, we adopt both pixel-level and perceptual metrics. While pixel-level metrics such as PSNR and SSIM have long been standard metrics in novel view synthesis, recent work~\cite{liang2025himor} has shown that these pixel-level metrics are misaligned with perceptual quality in monocular dynamic 3D reconstruction with view disparity, due to the inherent difficulty of predicting  `exact' position in this ill-posed setting. Therefore, we report both metric types together. For perceptual evaluation, we use LPIPS~\cite{zhang2018unreasonable} with AlexNet~\cite{krizhevsky2012imagenet} features and the CLIP~\cite{radford2021learning_clip}-based metrics proposed in~\cite{liang2025himor}:  CLIP-I, which measures CLIP similarity between ground-truth and rendered images, and CLIP-T, which measures temporal consistency via CLIP similarity between rendered frames with 5 frames apart.

\paragraph{iPhone360 dataset.}
We introduce the iPhone360 dataset to address limitations of existing benchmarks that cannot evaluate 360° dynamic reconstruction performance. 
Unlike prior datasets~\cite{gao2022dycheck,pumarola2021dnerf,park2021hypernerf,li2021nsff}, iPhone360 provides training images captured from a single handheld camera without camera teleportation, 
while synchronized test cameras are placed up to 70°–135° apart from the training view, enabling evaluation under extreme novel-view conditions. The large viewpoint disparity and realistic monocular capture setup make iPhone360 a suitable benchmark for evaluating the challenging 360° dynamic reconstruction.

\begin{figure}[t]
\begin{minipage}{0.48\linewidth}
\centering
\captionof{table}{Compact comparison across datasets. \textbf{\#Tr.}: number of training cameras, \textbf{Wild}: in-the-wild scenario, \textbf{Max $\Delta\theta$}: maximum angular difference between training and test views, \textbf{\boldmath $360^\circ$}: ability to evaluate reconstruction in $360^\circ$ manner.}
\vspace{7pt}
{\scriptsize
\setlength{\tabcolsep}{2pt}
\begin{tabular}{@{}lcccc@{}}
\toprule
\textbf{Dataset} & \textbf{\#Tr.} & \textbf{Wild} & \textbf{Max $\Delta\theta$} & \textbf{\boldmath $360^\circ$} \\
\midrule
D\text{-}NeRF~\cite{pumarola2021dnerf}    & $\sim$150 & \xmark & --          & \xmark \\
HyperNeRF~\cite{park2021hypernerf}        & 2         & \xmark & --          & \xmark \\
Nerfies~\cite{park2021nerfies}            & 2         & \xmark & --          & \xmark \\
NSFF~\cite{li2021nsff}                    & 24        & \xmark & --          & \xmark \\
iPhone~\cite{gao2022dycheck}              & 1         & \cmark & $<45^\circ$ & \xmark \\
\cmidrule(lr){1-5}
\textbf{iPhone360} & \textbf{1} & \textbf{\cmark} & \textbf{70°--135°} & \textbf{\cmark} \\
\bottomrule
\end{tabular}
}
\label{tab:datasets}
\end{minipage}
\hfill
\begin{minipage}{0.48\linewidth}
\centering
\includegraphics[width=\linewidth]{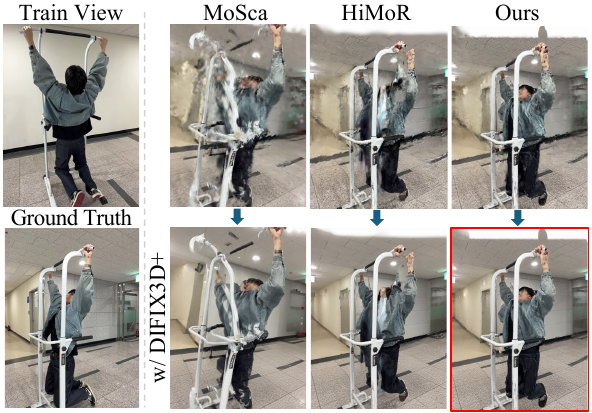}
\vspace{-15pt}
\caption{
\textbf{Results with Diffusion Prior.} Our method outperforms baselines when combined with DIFIX3D+~\cite{wu2025difix3d+}.
}
\label{fig:difix}
\end{minipage}
\end{figure}

The dataset consists of 6 dynamic scenes: \textit{block2}, \textit{goat}, \textit{jacket}, \textit{jelly}, \textit{pull-up},  and \textit{walk-around}. Each scene captures an object with realistic scenario. A compact comparison with existing dynamic datasets is shown in \cref{tab:datasets}. 
All sequences are captured using multiple iPhone devices, 
and follow the same acquisition framework as iPhone dataset. We use LiDAR metric depth to perform scale and shift correction on the estimated depth~\cite{chen2025videodepthanything, depth_anything_v2} results, and use camera parameters from the sensor as ground truth.

\begin{figure}[t]
    \centering
    \vspace{-10pt}
    \begin{minipage}[t]{0.50\linewidth}
        \centering
        \includegraphics[width=\linewidth]{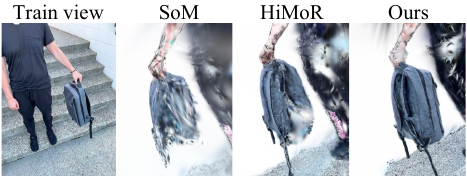}
        \vspace{-2em}
        \captionof{figure}{
        \textbf{iPhone dataset Qualitative result.} Backpack scene shows that our model faithfully reconstructs unseen backside regions, clearly outperforming prior methods.
        }
        \label{fig:backpack}
    \end{minipage}
    \hfill
    \begin{minipage}[t]{0.45\linewidth}
        \centering
        \vspace{-60pt}
        \captionof{table}{Pixel-level Evaluation on iPhone and iPhone360 Datasets. All training and evaluations are conducted at 1x resolution.}
        \label{tab:pixel_eval}
        \resizebox{\linewidth}{!}{%
        \begin{tabular}{lcccc}
        \toprule
        \multirow{2}{*}{} & \multicolumn{2}{c}{iPhone dataset} & \multicolumn{2}{c}{iPhone360 dataset} \\
        \cmidrule(lr){2-3} \cmidrule(lr){4-5}
         & PSNR$\uparrow$ & SSIM$\uparrow$ & PSNR$\uparrow$ & SSIM$\uparrow$ \\
        \midrule
        HiMoR & 16.2441 & \textbf{0.6329} & 10.9883 & 0.6854 \\
        Ours  & \textbf{16.3431} & 0.6309 & \textbf{11.3364} & \textbf{0.7012} \\
        \bottomrule
        \end{tabular}}
    \end{minipage}
    \vspace{-0.5em}
\end{figure}

In \cref{tab:iphone360_clip}, we report perceptual metrics on our iPhone360 dataset. Since rendered backgrounds in extreme novel view settings tend to contain large empty regions, direct image comparison becomes less meaningful. To address this, we define a bounding box centered on the ground-truth dynamic object mask expanded by a spatial margin, and apply it to both the ground-truth and foreground-only rendered results. All metrics are then computed within this region.
Our method achieves consistently higher scores on the iPhone360 dataset compared with baselines~\cite{liang2025himor, lei2025mosca}. As an additional reference alongside the CLIP-based metrics, we report pixel-level evaluation results in \cref{tab:pixel_eval}. The low absolute scores reflect the inherent difficulty of the task: achieving pixel-level accuracy in unseen regions under full extrapolation.

Furthermore, as shown \cref{fig:teaser} and \cref{fig:iphone360_qualitative}, our reconstructions maintain coherent geometry even from extreme viewpoints, demonstrating the effectiveness of the proposed method. \cref{fig:iphone360_qualitative}(b) demonstrates that our method achieves 360° object reconstruction from a single video.

\begin{table}[t]
\centering
\footnotesize
\caption{Quantitative results on iPhone360 dataset.}
\label{tab:iphone360_clip}
\resizebox{0.85\columnwidth}{!}{%
\begin{tabular}{lcccccccccc}
\toprule
& \multicolumn{3}{c}{Block2} & \multicolumn{3}{c}{Goat} & \multicolumn{3}{c}{Jacket} \\
\cmidrule(lr){2-4} \cmidrule(lr){5-7} \cmidrule(lr){8-10}
& CLIP-I$\uparrow$ & CLIP-T$\uparrow$ & LPIPS$\downarrow$ & CLIP-I$\uparrow$ & CLIP-T$\uparrow$ & LPIPS$\downarrow$ & CLIP-I$\uparrow$ & CLIP-T$\uparrow$ & LPIPS$\downarrow$ \\
\midrule
MoSca~\cite{lei2025mosca} & 0.7499 & 0.9513 & 0.5556 & 0.7379 & \textbf{0.9522} & 0.6332 & 0.7073 & 0.9440 & 0.5280 \\
HiMoR~\cite{liang2025himor} & 0.8422 & 0.9549 & 0.3122 & 0.8357 & 0.9426 & 0.3260 & 0.7041 & \textbf{0.9526} & 0.2942 \\
Ours  & \textbf{0.9021} & \textbf{0.9633} & \textbf{0.2569} & \textbf{0.8706} & 0.9345 & \textbf{0.2244} & \textbf{0.9147} & 0.9445 & \textbf{0.2747} \\
\midrule
& \multicolumn{3}{c}{Jelly} & \multicolumn{3}{c}{Pull-up} & \multicolumn{3}{c}{Walk-around} \\
\cmidrule(lr){2-4} \cmidrule(lr){5-7} \cmidrule(lr){8-10}
& CLIP-I$\uparrow$ & CLIP-T$\uparrow$ & LPIPS$\downarrow$ & CLIP-I$\uparrow$ & CLIP-T$\uparrow$ & LPIPS$\downarrow$ & CLIP-I$\uparrow$ & CLIP-T$\uparrow$ & LPIPS$\downarrow$ \\
\midrule
MoSca~\cite{lei2025mosca} & 0.6239 & \textbf{0.9649} & 0.7127 & 0.8081 & 0.9617 & 0.5241 & 0.7323 & \textbf{0.9639} & 0.7439 \\
HiMoR~\cite{liang2025himor} & 0.6839 & 0.9357 & 0.3539 & 0.8862 & 0.9786 & 0.2240 & 0.8333 & 0.9590 & 0.4014 \\
Ours  & \textbf{0.8359} & 0.9229 & \textbf{0.3538} & \textbf{0.8978} & \textbf{0.9827} & \textbf{0.2163} & \textbf{0.8718} & 0.9627 & \textbf{0.3938} \\
\bottomrule
\end{tabular}%
\vspace{-15pt}
}
\end{table}

\begin{table*}[t]
\centering
\setlength{\tabcolsep}{2pt}
\caption{Quantitative results of novel view synthesis on the iPhone dataset~\cite{gao2022dycheck}. Highlighted cells denote \firsttext{best}, \secondtext{second best}, \thirdtext{third best}.}
\resizebox{\textwidth}{!}{
\begin{tabular}{lccc|ccc|ccc}
\hline
\multirow{2}{*}{Method} & \multicolumn{3}{c|}{Apple} & \multicolumn{3}{c|}{Block} & \multicolumn{3}{c}{Paper-windmill} \\
\cmidrule(lr){2-4} \cmidrule(lr){5-7} \cmidrule(lr){8-10}
& CLIP-I$\uparrow$ & CLIP-T$\uparrow$ & LPIPS$\downarrow$ & CLIP-I$\uparrow$ & CLIP-T$\uparrow$ & LPIPS$\downarrow$ & CLIP-I$\uparrow$ & CLIP-T$\uparrow$ & LPIPS$\downarrow$ \\
\hline
T-NeRF~\cite{gao2022dycheck} & 0.8275 & 0.9729 & 0.6695 & \third{0.8873} & 0.9749 & \third{0.4729} & \second{0.9304} & \first{0.9858} & 0.4837 \\
HyperNeRF~\cite{park2021hypernerf} & \third{0.8314} & \second{0.9771} & 0.6626 & \second{0.8882} & \second{0.9756} & \second{0.4654} & 0.9218 & 0.9818 & 0.4319 \\
Deformable 3DGS~\cite{yang2024deformable} & 0.7822 & 0.9730 & 0.8558 & 0.8105 & 0.9720 & 0.7281 & 0.8652 & 0.9814 & 0.6411 \\
Marbles~\cite{stearns2024marbles} & 0.8055 & 0.9653 & 0.7025 & 0.8492 & 0.9648 & 0.5303 & 0.8610 & 0.9791 & 0.6280 \\
SoM~\cite{wang2025shape} & 0.8100 & 0.9721 & \third{0.6335} & 0.8658 & 0.9745 & 0.5083 & 0.9225 & 0.9833 & \third{0.3253} \\
HiMoR~\cite{liang2025himor} & \first{0.8798} & \third{0.9747} & \second{0.5926} & 0.8707 & \third{0.9750} & 0.5059 & \third{0.9275} & \third{0.9853} & \second{0.3216} \\
\hline
Ours & \second{0.8775} & \first{0.9794} & \first{0.5414} & \first{0.8933} & \first{0.9770} & \first{0.4105} & \first{0.9446} & \second{0.9854} & \first{0.2055} \\
\hline
\end{tabular}
}

\resizebox{\textwidth}{!}{
\begin{tabular}{lccc|ccc|ccc}
\hline
\multirow{2}{*}{Method} & \multicolumn{3}{c|}{Spin} & \multicolumn{3}{c|}{Teddy} & \multicolumn{3}{c}{\textbf{Mean}} \\
\cmidrule(lr){2-4} \cmidrule(lr){5-7} \cmidrule(lr){8-10}
& CLIP-I$\uparrow$ & CLIP-T$\uparrow$ & LPIPS$\downarrow$ & CLIP-I$\uparrow$ & CLIP-T$\uparrow$ & LPIPS$\downarrow$ & CLIP-I$\uparrow$ & CLIP-T$\uparrow$ & LPIPS$\downarrow$ \\
\hline
T-NeRF~\cite{gao2022dycheck} & 0.8328 & 0.9565 & 0.5714 & 0.8242 & 0.9541 & 0.6337 & 0.8604 & 0.9688 & 0.5662 \\
HyperNeRF~\cite{park2021hypernerf} & 0.8498 & 0.9594 & 0.4905 & \third{0.8836} & 0.9630 & 0.5801 & \third{0.8750} & 0.9714 & 0.5261 \\
Deformable 3DGS~\cite{yang2024deformable} & 0.7457 & \first{0.9712} & 0.5962 & 0.7791 & 0.9629 & 0.7601 & 0.7965 & \third{0.9721} & 0.7163 \\
Marbles~\cite{stearns2024marbles} & 0.8272 & 0.9527 & 0.5761 & 0.8097 & 0.9606 & 0.6671 & 0.8305 & 0.9645 & 0.6208 \\
SoM~\cite{wang2025shape} & \third{0.8510} & 0.9585 & \third{0.3832} & 0.8521 & \third{0.9676} & \third{0.5630} & 0.8603 & 0.9712 & \third{0.4827} \\
HiMoR~\cite{liang2025himor} & \second{0.8853} & \second{0.9658} & \second{0.3696} & \first{0.8902} & \first{0.9744} & \second{0.5296} & \second{0.8907} & \second{0.9750} & \second{0.4639} \\
\hline
Ours & \first{0.9029} & \third{0.9626} & \first{0.2957} & \second{0.8891} & \second{0.9726} & \first{0.4856} & \first{0.9015} & \first{0.9754} & \first{0.3877} \\
\hline
\label{tab:iphone_clip}
\end{tabular}
}
\end{table*}

\begin{figure}[t]
    \centering
    \includegraphics[width=\linewidth]{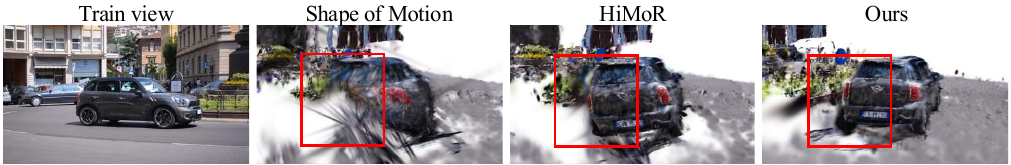}
    \caption{
    \textbf{DAVIS dataset Qualitative Comparison}. Results show that our model maintains coherent geometry and recovers occluded parts more reliably than previous approaches.
    }
    \label{fig:davis}
\end{figure}
\begin{figure}[t]
    \centering
    \includegraphics[width=\linewidth]{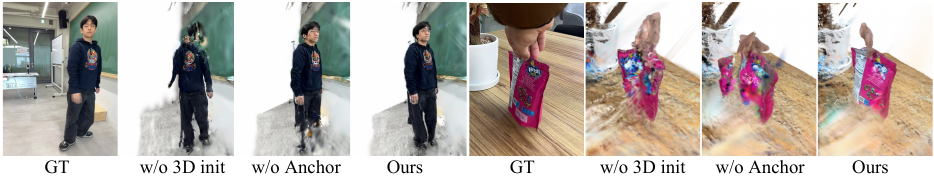}
    \caption{
    \textbf{Ablations.} Qualitative results for different initialization methods. `w/o 3D init' directly unprojects 2D tracks, `w/o Anchor' leverages naive 3D point tracking model~\cite{tapip3d}, and Ours uses AnchorTAP3D at initialization. The left and right scenes correspond to the \textit{walk-around} and \textit{jelly} sequences from iPhone360, respectively.
    }
    \label{fig:ablation}
\end{figure}

\paragraph{iPhone dataset.}
The iPhone~\cite{gao2022dycheck} dataset contains training videos captured while moving between two test cameras, designed to evaluate dynamic reconstruction models in a monocular setting.
Following HiMoR~\cite{liang2025himor}, we report CLIP-I, CLIP-T, and LPIPS~\cite{zhang2018unreasonable} scores on \cref{tab:iphone_clip}. Across these metrics, our method shows consistent improvements over previous models. We compare our model with NeRF-based~\cite{gao2022dycheck,park2021hypernerf} and 3DGS-based~\cite{stearns2024marbles,wang2025shape,liang2025himor,yang2024deformable} approaches across five scenes. Our method outperforms on most scenes. The qualitative results on the \textit{spin} scene are included in \cref{fig:teaser}. Additionally, \cref{fig:backpack} presents extreme-viewpoint renderings for the \textit{backpack} scene, which lacks test ground truth, showing that our model reconstructs unseen regions with substantially higher fidelity than competing methods~\cite{wang2025shape, liang2025himor}.

\paragraph{DAVIS dataset.}
We evaluated our model on scenes from the DAVIS~\cite{ponttuset20182017davischallengevideo} dataset, which captures fast-moving objects but lacks ground-truth camera parameters and depth maps, making them closer to real-world conditions. As shown in \cref{fig:davis}, our method better preserves geometry and reconstructs occluded regions compared to previous models under in-the-wild scenarios.

\subsection{Ablations}
We conducted a qualitative ablation study on the \textit{walk-around} and \textit{jelly} scenes of the iPhone360 dataset to demonstrate the advantage of AnchorTAP3D over baselines using naive 2D and 3D tracking. As shown in \cref{fig:ablation}, our method achieves better 360° reconstruction, reconstructing the geometry of occluded regions at extreme viewpoints, compared to `w/o 3D init' and `w/o Anchor'. 

The `w/o 3D init' variant follows prior models~\cite{wang2025shape, liang2025himor} in initializing with unprojected 2D tracks, and fails to preserve geometry in occluded regions. 
Using a naive 3D point tracking model~\cite{tapip3d} `w/o Anchor' yields better performance than `w/o 3D init' in \textit{walk-around} by partially recovering geometry in occluded regions, but in the \textit{jelly}, it completely failed to reconstruct occluded area. This indicates that 3D tracking becomes unreliable in long sequences without anchor guidance. Finally, Ours preserves the overall object shape across both scenes. These results demonstrate the effectiveness of our method in 4D reconstruction over both 2D track-based and naive 3D tracking approaches.
\section{Limitations and Conclusion}
\label{sec:conclusion}

\paragraph{Limitations.}
While our method demonstrates improved 360° object reconstruction compared to existing approaches, several limitations remain. While our AnchorTAP3D improves significantly upon naive applications of off-the-shelf 2D and 3D tracking models, the overall performance still depends on the capability of pretrained models. 
Furthermore, our method cannot synthesize extreme viewpoint's background regions if it is invisible in the input video. We believe this can be addressed in future work through integration with diffusion priors. As shown in \cref{fig:difix}, our method, with its better-preserved geometry, provides a more suitable starting point for diffusion model than existing approaches. Our method also still struggles with pixel-level metrics such as PSNR and SSIM under the inherently challenging large extrapolation settings of iPhone360, and we leave further improvement of pixel-level reconstruction accuracy to future work.

\paragraph{Conclusion.}
We present 4DGS360, a novel framework that enables extreme novel-view synthesis from monocular videos. Unlike prior methods that rely on pretrained 2D-based features and suffer from 3D ambiguity, we introduce AnchorTAP3D, an advanced 3D tracking model that provides reliable geometry-aware initialization. This design extends the effectiveness of optimization strategies, allowing consistent 360° reconstruction of dynamic objects.
Furthermore, we identify limitations in current datasets and propose a new benchmark, iPhone360, which includes dynamic objects exhibiting diverse motion patterns with extreme viewpoint validations. This dataset enables more realistic evaluation of how well reconstruction models generalize to challenging real-world conditions.


\section*{Acknowledgements}
This work was supported by the Korean Government through the grants from IITP (RS-2021-II211343 and RS-2025-25442338) and COMPA (RS-2026-25537086).

%
%
\bibliographystyle{splncs04}
\bibliography{main}

@String(CVPR= {IEEE Conf. Comput. Vis. Pattern Recog.})

@String(ICCV= {Int. Conf. Comput. Vis.})

@String(ECCV= {Eur. Conf. Comput. Vis.})

@String(NIPS= {Adv. Neural Inform. Process. Syst.})

@String(TOG= {ACM Trans. Graph.})

@String(ACCV  = {ACCV})

@String(AAAI = {AAAI})

@String(CVPR  = {Conf. Comput. Vis. Pattern Recog.})

@String(ICCV  = {ICCV})

@String(ECCV  = {ECCV})

@String(NIPS  = {NeurIPS})

@String(TOG   = {ACM TOG})

@inproceedings{gao2022dycheck,
title={Monocular dynamic view synthesis: A reality check},
author={Gao, Hang and Li, Ruilong and Tulsiani, Shubham and Russell, Bryan and Kanazawa, Angjoo},
booktitle={NeurIPS},
year={2022},
}

@article{park2021hypernerf,
  author = {Park, Keunhong and Sinha, Utkarsh and Hedman, Peter and Barron, Jonathan T. and Bouaziz, Sofien and Goldman, Dan B and Martin-Brualla, Ricardo and Seitz, Steven M.},
  title = {HyperNeRF: A Higher-Dimensional Representation for Topologically Varying Neural Radiance Fields},
  journal = {ACM Trans. Graph.},
  issue_date = {December 2021},
  publisher = {ACM},
  volume = {40},
  number = {6},
  month = {dec},
  year = {2021},
  articleno = {238},
}

@inproceedings{yang2024deformable,
  title={Deformable 3d gaussians for high-fidelity monocular dynamic scene reconstruction},
  author={Yang, Ziyi and Gao, Xinyu and Zhou, Wen and Jiao, Shaohui and Zhang, Yuqing and Jin, Xiaogang},
  booktitle={Proceedings of the IEEE/CVF conference on computer vision and pattern recognition},
  pages={20331--20341},
  year={2024}
}

@inproceedings{stearns2024marbles,
  title={Dynamic gaussian marbles for novel view synthesis of casual monocular videos},
  author={Stearns, Colton and Harley, Adam and Uy, Mikaela and Dubost, Florian and Tombari, Federico and Wetzstein, Gordon and Guibas, Leonidas},
  booktitle={SIGGRAPH Asia 2024 Conference Papers},
  pages={1--11},
  year={2024}
}

@inproceedings{wang2025shape,
  title={Shape of motion: 4d reconstruction from a single video},
  author={Wang, Qianqian and Ye, Vickie and Gao, Hang and Zeng, Weijia and Austin, Jake and Li, Zhengqi and Kanazawa, Angjoo},
  booktitle={Proceedings of the IEEE/CVF International Conference on Computer Vision},
  pages={9660--9672},
  year={2025}
}

@inproceedings{liang2025himor,
  title={HiMoR: Monocular Deformable Gaussian Reconstruction with Hierarchical Motion Representation},
  author={Liang, Yiming and Xu, Tianhan and Kikuchi, Yuta},
  booktitle={Proceedings of the Computer Vision and Pattern Recognition Conference},
  pages={886--895},
  year={2025}
}

@inproceedings{pumarola2021dnerf,
  title={D-nerf: Neural radiance fields for dynamic scenes},
  author={Pumarola, Albert and Corona, Enric and Pons-Moll, Gerard and Moreno-Noguer, Francesc},
  booktitle={Proceedings of the IEEE/CVF conference on computer vision and pattern recognition},
  pages={10318--10327},
  year={2021}
}

@inproceedings{li2021nsff,
  title={Neural scene flow fields for space-time view synthesis of dynamic scenes},
  author={Li, Zhengqi and Niklaus, Simon and Snavely, Noah and Wang, Oliver},
  booktitle={Proceedings of the IEEE/CVF conference on computer vision and pattern recognition},
  pages={6498--6508},
  year={2021}
}

@inproceedings{park2021nerfies,
  title={Nerfies: Deformable neural radiance fields},
  author={Park, Keunhong and Sinha, Utkarsh and Barron, Jonathan T and Bouaziz, Sofien and Goldman, Dan B and Seitz, Steven M and Martin-Brualla, Ricardo},
  booktitle={Proceedings of the IEEE/CVF international conference on computer vision},
  pages={5865--5874},
  year={2021}
}

@article{kerbl20233d,
  title={3D Gaussian splatting for real-time radiance field rendering.},
  author={Kerbl, Bernhard and Kopanas, Georgios and Leimk{\"u}hler, Thomas and Drettakis, George},
  journal={ACM Trans. Graph.},
  volume={42},
  number={4},
  pages={139--1},
  year={2023}
}

@article{lee2023fast,
  title={Fast View Synthesis of Casual Videos with Soup-of-Planes},
  author={Lee, Yao-Chih and Zhang, Zhoutong and Blackburn-Matzen, Kevin and Niklaus, Simon and Zhang, Jianming and Huang, Jia-Bin and Liu, Feng},
  journal={arXiv preprint arXiv:2312.02135},
  year={2023}
}

@inproceedings{liu2023robust,
  title={Robust dynamic radiance fields},
  author={Liu, Yu-Lun and Gao, Chen and Meuleman, Andreas and Tseng, Hung-Yu and Saraf, Ayush and Kim, Changil and Chuang, Yung-Yu and Kopf, Johannes and Huang, Jia-Bin},
  booktitle={Proceedings of the IEEE/CVF Conference on Computer Vision and Pattern Recognition},
  pages={13--23},
  year={2023}
}

@inproceedings{lei2025mosca,
  title={Mosca: Dynamic gaussian fusion from casual videos via 4d motion scaffolds},
  author={Lei, Jiahui and Weng, Yijia and Harley, Adam W and Guibas, Leonidas and Daniilidis, Kostas},
  booktitle={Proceedings of the Computer Vision and Pattern Recognition Conference},
  pages={6165--6177},
  year={2025}
}

@inproceedings{doersch2023tapir,
  title={Tapir: Tracking any point with per-frame initialization and temporal refinement},
  author={Doersch, Carl and Yang, Yi and Vecerik, Mel and Gokay, Dilara and Gupta, Ankush and Aytar, Yusuf and Carreira, Joao and Zisserman, Andrew},
  booktitle={Proceedings of the IEEE/CVF International Conference on Computer Vision},
  pages={10061--10072},
  year={2023}
}

@InProceedings{Doersch_2024_bootstap,
    author    = {Doersch, Carl and Luc, Pauline and Yang, Yi and Gokay, Dilara and Koppula, Skanda and Gupta, Ankush and Heyward, Joseph and Rocco, Ignacio and Goroshin, Ross and Carreira, Jo\~ao and Zisserman, Andrew},
    title     = {BootsTAP: Bootstrapped Training for Tracking-Any-Point},
    booktitle = {Proceedings of the Asian Conference on Computer Vision (ACCV)},
    month     = {December},
    year      = {2024},
    pages     = {3257-3274}
}

@article{chu2024dreamscene4d,
  title={Dreamscene4d: Dynamic multi-object scene generation from monocular videos},
  author={Chu, Wen-Hsuan and Ke, Lei and Fragkiadaki, Katerina},
  journal={Advances in Neural Information Processing Systems},
  volume={37},
  pages={96181--96206},
  year={2024}
}

@inproceedings{rombach2022high,
  title={High-resolution image synthesis with latent diffusion models},
  author={Rombach, Robin and Blattmann, Andreas and Lorenz, Dominik and Esser, Patrick and Ommer, Bj{\"o}rn},
  booktitle={Proceedings of the IEEE/CVF conference on computer vision and pattern recognition},
  pages={10684--10695},
  year={2022}
}

@inproceedings{radford2021learning_clip,
  title={Learning transferable visual models from natural language supervision},
  author={Radford, Alec and Kim, Jong Wook and Hallacy, Chris and Ramesh, Aditya and Goh, Gabriel and Agarwal, Sandhini and Sastry, Girish and Askell, Amanda and Mishkin, Pamela and Clark, Jack and others},
  booktitle={International conference on machine learning},
  pages={8748--8763},
  year={2021},
  organization={PmLR}
}

@article{mildenhall2021nerf,
  title={Nerf: Representing scenes as neural radiance fields for view synthesis},
  author={Mildenhall, Ben and Srinivasan, Pratul P and Tancik, Matthew and Barron, Jonathan T and Ramamoorthi, Ravi and Ng, Ren},
  journal={Communications of the ACM},
  volume={65},
  number={1},
  pages={99--106},
  year={2021},
  publisher={ACM New York, NY, USA}
}

@inproceedings{barron2021mip,
  title={Mip-nerf: A multiscale representation for anti-aliasing neural radiance fields},
  author={Barron, Jonathan T and Mildenhall, Ben and Tancik, Matthew and Hedman, Peter and Martin-Brualla, Ricardo and Srinivasan, Pratul P},
  booktitle={Proceedings of the IEEE/CVF international conference on computer vision},
  pages={5855--5864},
  year={2021}
}

@inproceedings{barron2022mip,
  title={Mip-nerf 360: Unbounded anti-aliased neural radiance fields},
  author={Barron, Jonathan T and Mildenhall, Ben and Verbin, Dor and Srinivasan, Pratul P and Hedman, Peter},
  booktitle={Proceedings of the IEEE/CVF conference on computer vision and pattern recognition},
  pages={5470--5479},
  year={2022}
}

@inproceedings{seo2023flipnerf,
  title={Flipnerf: Flipped reflection rays for few-shot novel view synthesis},
  author={Seo, Seunghyeon and Chang, Yeonjin and Kwak, Nojun},
  booktitle={Proceedings of the IEEE/CVF International Conference on Computer Vision},
  pages={22883--22893},
  year={2023}
}

@inproceedings{wang2022clip,
  title={Clip-nerf: Text-and-image driven manipulation of neural radiance fields},
  author={Wang, Can and Chai, Menglei and He, Mingming and Chen, Dongdong and Liao, Jing},
  booktitle={Proceedings of the IEEE/CVF conference on computer vision and pattern recognition},
  pages={3835--3844},
  year={2022}
}

@article{verbin2024ref,
  title={Ref-nerf: Structured view-dependent appearance for neural radiance fields},
  author={Verbin, Dor and Hedman, Peter and Mildenhall, Ben and Zickler, Todd and Barron, Jonathan T and Srinivasan, Pratul P},
  journal={IEEE Transactions on Pattern Analysis and Machine Intelligence},
  year={2024},
  publisher={IEEE}
}

@inproceedings{yu2021pixelnerf,
  title={pixelnerf: Neural radiance fields from one or few images},
  author={Yu, Alex and Ye, Vickie and Tancik, Matthew and Kanazawa, Angjoo},
  booktitle={Proceedings of the IEEE/CVF conference on computer vision and pattern recognition},
  pages={4578--4587},
  year={2021}
}

@article{lee2025divcon,
  title={DivCon-NeRF: Generating Augmented Rays with Diversity and Consistency for Few-shot View Synthesis},
  author={Lee, Ingyun and Jang, Jae Won and Seo, Seunghyeon and Kwak, Nojun},
  journal={arXiv preprint arXiv:2503.12947},
  year={2025}
}

@inproceedings{barron2023zip,
  title={Zip-nerf: Anti-aliased grid-based neural radiance fields},
  author={Barron, Jonathan T and Mildenhall, Ben and Verbin, Dor and Srinivasan, Pratul P and Hedman, Peter},
  booktitle={Proceedings of the IEEE/CVF International Conference on Computer Vision},
  pages={19697--19705},
  year={2023}
}

@article{duckworth2024smerf,
  title={SMERF: Streamable memory efficient radiance fields for real-time large-scene exploration},
  author={Duckworth, Daniel and Hedman, Peter and Reiser, Christian and Zhizhin, Peter and Thibert, Jean-Fran{\c{c}}ois and Lu{\v{c}}i{\'c}, Mario and Szeliski, Richard and Barron, Jonathan T},
  journal={ACM Transactions on Graphics (TOG)},
  volume={43},
  number={4},
  pages={1--13},
  year={2024},
  publisher={ACM New York, NY, USA}
}

@inproceedings{wu20244dgs,
  title={4d gaussian splatting for real-time dynamic scene rendering},
  author={Wu, Guanjun and Yi, Taoran and Fang, Jiemin and Xie, Lingxi and Zhang, Xiaopeng and Wei, Wei and Liu, Wenyu and Tian, Qi and Wang, Xinggang},
  booktitle={Proceedings of the IEEE/CVF conference on computer vision and pattern recognition},
  pages={20310--20320},
  year={2024}
}

@inproceedings{luiten2024dynamic,
  title={Dynamic 3d gaussians: Tracking by persistent dynamic view synthesis},
  author={Luiten, Jonathon and Kopanas, Georgios and Leibe, Bastian and Ramanan, Deva},
  booktitle={2024 International Conference on 3D Vision (3DV)},
  pages={800--809},
  year={2024},
  organization={IEEE}
}

@inproceedings{wu20254dfly,
  title={4D-Fly: Fast 4D Reconstruction from a Single Monocular Video},
  author={Wu, Diankun and Liu, Fangfu and Hung, Yi-Hsin and Qian, Yue and Zhan, Xiaohang and Duan, Yueqi},
  booktitle={Proceedings of the Computer Vision and Pattern Recognition Conference},
  pages={16663--16673},
  year={2025}
}

@inproceedings{lu2024scaffold,
  title={Scaffold-gs: Structured 3d gaussians for view-adaptive rendering},
  author={Lu, Tao and Yu, Mulin and Xu, Linning and Xiangli, Yuanbo and Wang, Limin and Lin, Dahua and Dai, Bo},
  booktitle={Proceedings of the IEEE/CVF Conference on Computer Vision and Pattern Recognition},
  pages={20654--20664},
  year={2024}
}

@inproceedings{mallick2024taming,
  title={Taming 3dgs: High-quality radiance fields with limited resources},
  author={Mallick, Saswat Subhajyoti and Goel, Rahul and Kerbl, Bernhard and Steinberger, Markus and Carrasco, Francisco Vicente and De La Torre, Fernando},
  booktitle={SIGGRAPH Asia 2024 Conference Papers},
  pages={1--11},
  year={2024}
}

@inproceedings{Shen2024SuperGaussian,
  title = {SuperGaussian: Repurposing Video Models for 3D Super Resolution},
  author = {Shen, Yuan and Ceylan, Duygu and Guerrero, Paul and Xu, Zexiang and Mitra, {Niloy J.} and Wang, Shenlong and Fr{\"u}hst{\"u}ck, Anna},
  booktitle = {European Conference on Computer Vision (ECCV)},
  year = {2024},
}

@InProceedings{Yu2024MipSplatting,
    author    = {Yu, Zehao and Chen, Anpei and Huang, Binbin and Sattler, Torsten and Geiger, Andreas},
    title     = {Mip-Splatting: Alias-free 3D Gaussian Splatting},
    booktitle = {Proceedings of the IEEE/CVF Conference on Computer Vision and Pattern Recognition (CVPR)},
    month     = {June},
    year      = {2024},
    pages     = {19447-19456}
}

@article{Chen_deblurgs2024,
   author       = {Wenbo, Chen and Ligang, Liu},
   title        = {Deblur-GS: 3D Gaussian Splatting from Camera Motion Blurred Images},
   journal      = {Proc. ACM Comput. Graph. Interact. Tech. (Proceedings of I3D 2024)},
   year         = {2024},
   volume       = {7},
   number       = {1},
   numpages     = {13},
   location     = {Philadelphia, PA, USA},
   url          = {http://doi.acm.org/10.1145/3651301},
   doi          = {10.1145/3651301},
   publisher    = {ACM Press},
   address      = {New York, NY, USA},
}

@inproceedings{zhao2024badgaussians,
    author = {Zhao, Lingzhe and Wang, Peng and Liu, Peidong},
    title = {Bad-gaussians: Bundle adjusted deblur gaussian splatting},
    booktitle = {European Conference on Computer Vision (ECCV)},
    year = {2024}
}

@inproceedings{chen2024mvsplat,
 author = {Chen, Yuedong and Zheng, Chuanxia and Xu, Haofei and Zhuang, Bohan and Vedaldi, Andrea and Cham, Tat-Jen and Cai, Jianfei},
 booktitle = {Advances in Neural Information Processing Systems},
 doi = {10.52202/079017-3399},
 editor = {A. Globerson and L. Mackey and D. Belgrave and A. Fan and U. Paquet and J. Tomczak and C. Zhang},
 pages = {107064--107086},
 publisher = {Curran Associates, Inc.},
 title = {MVSplat360: Feed-Forward 360 Scene Synthesis from Sparse Views},
 url = {https://proceedings.neurips.cc/paper_files/paper/2024/file/c196239c5f9481e0db2755f31fe4585f-Paper-Conference.pdf},
 volume = {37},
 year = {2024}
}

@inproceedings{liu2025mvsgaussian,
    title={MVSGaussian: Fast Generalizable Gaussian Splatting Reconstruction from Multi-View Stereo},
    author={Liu, Tianqi and Wang, Guangcong and Hu, Shoukang and Shen, Liao and Ye, Xinyi and Zang, Yuhang and Cao, Zhiguo and Li, Wei and Liu, Ziwei},
    booktitle={European Conference on Computer Vision},
    pages={37--53},
    year={2025},
    organization={Springer}
}

@inproceedings{paliwal2024coherentgs,
    title={CoherentGS: Sparse Novel View Synthesis with Coherent 3D Gaussians},
    author={Paliwal, Avinash and Ye, Wei and Xiong, Jinhui and Kotovenko, Dmytro and Ranjan, Rakesh and Chandra, Vikas
            and Kalantari, Nima Khademi},
    booktitle={European Conference on Computer Vision},
    pages={19--37},
    year={2024},
    organization={Springer}
}

@inproceedings{charatan23pixelsplat,
      title={pixelSplat: 3D Gaussian Splats from Image Pairs for Scalable Generalizable 3D Reconstruction},
      author={David Charatan and Sizhe Li and Andrea Tagliasacchi and Vincent Sitzmann},
      year={2024},
      booktitle={CVPR},
}

@misc{hu2024evagaussian3dgaussianbasedrealtime,
    title={EVA-Gaussian: 3D Gaussian-based Real-time Human Novel View Synthesis under Diverse Camera Settings}, 
    author={Yingdong Hu and Zhening Liu and Jiawei Shao and Zehong Lin and Jun Zhang},
    year={2024},
    eprint={2410.01425},
    archivePrefix={arXiv},
    primaryClass={cs.CV},
    url={https://arxiv.org/abs/2410.01425}, 
}

@misc{yang2023track,
      title={Track Anything: Segment Anything Meets Videos}, 
      author={Jinyu Yang and Mingqi Gao and Zhe Li and Shang Gao and Fangjing Wang and Feng Zheng},
      year={2023},
      eprint={2304.11968},
      archivePrefix={arXiv},
      primaryClass={cs.CV}
}

@article{depth_anything_v2,
  title={Depth anything v2},
  author={Yang, Lihe and Kang, Bingyi and Huang, Zilong and Zhao, Zhen and Xu, Xiaogang and Feng, Jiashi and Zhao, Hengshuang},
  journal={Advances in Neural Information Processing Systems},
  volume={37},
  pages={21875--21911},
  year={2024}
}

@inproceedings{depthanything,
      title={Depth Anything: Unleashing the Power of Large-Scale Unlabeled Data}, 
      author={Yang, Lihe and Kang, Bingyi and Huang, Zilong and Xu, Xiaogang and Feng, Jiashi and Zhao, Hengshuang},
      booktitle={CVPR},
      year={2024}
}

@InProceedings{Wang_2024_CVPR,
    author    = {Wang, Shuzhe and Leroy, Vincent and Cabon, Yohann and Chidlovskii, Boris and Revaud, Jerome},
    title     = {DUSt3R: Geometric 3D Vision Made Easy},
    booktitle = {Proceedings of the IEEE/CVF Conference on Computer Vision and Pattern Recognition (CVPR)},
    month     = {June},
    year      = {2024},
    pages     = {20697-20709}
}

@inproceedings{10.5555/3540261.3541527,
author = {Teed, Zachary and Deng, Jia},
title = {DROID-SLAM: deep visual SLAM for monocular, stereo, and RGB-D cameras},
year = {2021},
isbn = {9781713845393},
publisher = {Curran Associates Inc.},
address = {Red Hook, NY, USA},
booktitle = {Proceedings of the 35th International Conference on Neural Information Processing Systems},
articleno = {1266},
numpages = {12},
series = {NIPS '21}
}

@InProceedings{Kirillov_2023_ICCV,
    author    = {Kirillov, Alexander and Mintun, Eric and Ravi, Nikhila and Mao, Hanzi and Rolland, Chloe and Gustafson, Laura and Xiao, Tete and Whitehead, Spencer and Berg, Alexander C. and Lo, Wan-Yen and Dollar, Piotr and Girshick, Ross},
    title     = {Segment Anything},
    booktitle = {Proceedings of the IEEE/CVF International Conference on Computer Vision (ICCV)},
    month     = {October},
    year      = {2023},
    pages     = {4015-4026}
}

@inproceedings{Harley2022ParticleVR,
  title={Particle Video Revisited: Tracking Through Occlusions Using Point Trajectories},
  author={Adam W. Harley and Zhaoyuan Fang and Katerina Fragkiadaki},
  booktitle={European Conference on Computer Vision},
  year={2022},
  url={https://api.semanticscholar.org/CorpusID:248069518}
}

@inproceedings{karaev23cotracker,
  title     = {CoTracker: It is Better to Track Together},
  author    = {Nikita Karaev and Ignacio Rocco and Benjamin Graham and Natalia Neverova and Andrea Vedaldi and Christian Rupprecht},
  booktitle = {Proc. {ECCV}},
  year      = {2024}
}

@inproceedings{karaev24cotracker3,
  title     = {CoTracker3: Simpler and Better Point Tracking by Pseudo-Labelling Real Videos},
  author    = {Nikita Karaev and Iurii Makarov and Jianyuan Wang and Natalia Neverova and Andrea Vedaldi and Christian Rupprecht},
  booktitle = {Proc. {arXiv:2410.11831}},
  year      = {2024}
}

@article{tapip3d,
  title={TAPIP3D: Tracking Any Point in Persistent 3D Geometry},
  author={Zhang, Bowei and Ke, Lei and Harley, Adam W and Fragkiadaki, Katerina},
  journal={arXiv preprint arXiv:2504.14717},
  year={2025}
}

@article{Schonemann_1966, title={A Generalized Solution of the Orthogonal Procrustes Problem}, volume={31}, DOI={10.1007/BF02289451}, number={1}, journal={Psychometrika}, author={Schönemann, Peter H.}, year={1966}, pages={1–10}}

@inproceedings{arap,
author = {Sorkine, Olga and Alexa, Marc},
title = {As-rigid-as-possible surface modeling},
year = {2007},
isbn = {9783905673463},
publisher = {Eurographics Association},
address = {Goslar, DEU},
booktitle = {Proceedings of the Fifth Eurographics Symposium on Geometry Processing},
pages = {109–116},
numpages = {8},
location = {Barcelona, Spain},
series = {SGP '07}
}

@article{article_ssim,
  title={Image quality assessment: From error visibility to structural similarity},
  author={Wang, Zhou and Bovik, Alan and Sheikh, Hamid and Simoncelli, Eero},
  journal={IEEE Transactions on Image Processing},
  volume={13},
  number={4},
  pages={600--612},
  year={2004},
  publisher={IEEE}
}

@inproceedings{zhang2018unreasonable,
  title={The unreasonable effectiveness of deep features as a perceptual metric},
  author={Zhang, Richard and Isola, Phillip and Efros, Alexei A and Shechtman, Eli and Wang, Oliver},
  booktitle=CVPR,
  pages={586--595},
  year={2018}
}

@article{krizhevsky2012imagenet,
  title={Imagenet classification with deep convolutional neural networks},
  author={Krizhevsky, Alex and Sutskever, Ilya and Hinton, Geoffrey E},
  journal=NIPS,
  volume={25},
  year={2012}
}

@misc{ponttuset20182017davischallengevideo,
      title={The 2017 DAVIS Challenge on Video Object Segmentation}, 
      author={Jordi Pont-Tuset and Federico Perazzi and Sergi Caelles and Pablo Arbeláez and Alex Sorkine-Hornung and Luc Van Gool},
      year={2018},
      eprint={1704.00675},
      archivePrefix={arXiv},
      primaryClass={cs.CV},
      url={https://arxiv.org/abs/1704.00675}, 
}

@article{Kingma2014AdamAM,
  title={Adam: A method for stochastic optimization},
  author={Diederik P. Kingma and Jimmy Ba},
  journal={CoRR},
  year={2014},
  volume={abs/1412.6980},
  url={https://api.semanticscholar.org/CorpusID:6628106}
}

@inproceedings{wu2025difix3d+,
  title={Difix3d+: Improving 3d reconstructions with single-step diffusion models},
  author={Wu, Jay Zhangjie and Zhang, Yuxuan and Turki, Haithem and Ren, Xuanchi and Gao, Jun and Shou, Mike Zheng and Fidler, Sanja and Gojcic, Zan and Ling, Huan},
  booktitle=CVPR,
  pages={26024--26035},
  year={2025}
}

@inproceedings{xiao2025spatialtrackerv2,
      title={SpatialTrackerV2: 3D Point Tracking Made Easy}, 
      author={Yuxi Xiao and Jianyuan Wang and Nan Xue and Nikita Karaev and Yuri Makarov and Bingyi Kang and Xing Zhu and Hujun Bao and Yujun Shen and Xiaowei Zhou},
      year={2025},
      booktitle={Proceedings of the IEEE/CVF International Conference on Computer Vision},
      url={https://arxiv.org/abs/2507.12462}, 
}

@misc{sam3dteam2025sam3d3dfyimages,
      title={SAM 3D: 3Dfy Anything in Images}, 
      author={SAM 3D Team and Xingyu Chen and Fu-Jen Chu and Pierre Gleize and Kevin J Liang and Alexander Sax and Hao Tang and Weiyao Wang and Michelle Guo and Thibaut Hardin and Xiang Li and Aohan Lin and Jiawei Liu and Ziqi Ma and Anushka Sagar and Bowen Song and Xiaodong Wang and Jianing Yang and Bowen Zhang and Piotr Dollár and Georgia Gkioxari and Matt Feiszli and Jitendra Malik},
      year={2025},
      eprint={2511.16624},
      archivePrefix={arXiv},
      primaryClass={cs.CV},
      url={https://arxiv.org/abs/2511.16624}, 
}

@article{kim20244d,
  title={4d gaussian splatting in the wild with uncertainty-aware regularization},
  author={Kim, Mijeong and Lim, Jongwoo and Han, Bohyung},
  journal={Advances in Neural Information Processing Systems},
  volume={37},
  pages={129209--129226},
  year={2024}
}

@inproceedings{liu2025modgs,
  title={Modgs: Dynamic gaussian splatting from casually-captured monocular videos with depth priors},
  author={Liu, Qingming and Liu, Yuan and Wang, Jiepeng and Lyu, Xianqiang and Wang, Peng and Wang, Wenping and Hou, Junhui},
  booktitle={International Conference on Learning Representations},
  volume={2025},
  pages={97048--97074},
  year={2025}
}

@inproceedings{wang2025gflow,
  title={Gflow: Recovering 4d world from monocular video},
  author={Wang, Shizun and Yang, Xingyi and Shen, Qiuhong and Jiang, Zhenxiang and Wang, Xinchao},
  booktitle={Proceedings of the AAAI Conference on Artificial Intelligence},
  volume={39},
  pages={7862--7870},
  year={2025}
}

@inproceedings{chen2025videodepthanything,
  title={Video depth anything: Consistent depth estimation for super-long videos},
  author={Chen, Sili and Guo, Hengkai and Zhu, Shengnan and Zhang, Feihu and Huang, Zilong and Feng, Jiashi and Kang, Bingyi},
  booktitle={Proceedings of the Computer Vision and Pattern Recognition Conference},
  pages={22831--22840},
  year={2025}
}

@article{wang2026moge,
  title={Moge-2: Accurate monocular geometry with metric scale and sharp details},
  author={Wang, Ruicheng and Xu, Sicheng and Dong, Yue and Deng, Yu and Xiang, Jianfeng and Lv, Zelong and Sun, Guangzhong and Tong, Xin and Yang, Jiaolong},
  journal={Advances in Neural Information Processing Systems},
  volume={38},
  pages={35928--35959},
  year={2026}
}

\clearpage
\setcounter{page}{1}

\appendix

\begin{center}
    \Large
    \textbf{4DGS360: 360° Gaussian Reconstruction of Dynamic Objects from a Single Video} \
    \vspace{0.7em}
    {\large Supplementary Material}
    \vspace{1.0em}
\end{center}

\noindent
In this appendix, we present further implementation details, more results, and ablations. The additional material is structured as follows:
\begin{itemize}[label=\textbullet, itemsep=7pt]
    \item Sec. A: Additional Details
        \begin{itemize}[label=$\circ$, itemsep=3pt, topsep=4pt]
        \item Sec. A.1: AnchorTAP3D details
        \item Sec. A.2: Additional initialization details
        \item Sec. A.3: Optimization details
        \end{itemize}
    \item Sec. B: Additional Results
        \begin{itemize}[label=$\circ$, itemsep=3pt, topsep=4pt]
        \item Sec. B.1: Evaluation details
        \item Sec. B.2: iPhone360 dataset
        \item Sec. B.3: iPhone dataset
        \end{itemize}
    \item Sec. C: Additional Ablation
    \item Sec. D: iPhone360 Dataset Overview
\end{itemize}

\section{Additional Details}
\label{sec:add_details}
\subsection{AnchorTAP3D details}
\label{subsec:anchortap3d_details}
AnchorTAP3D employs BootsTAP~\cite{Doersch_2024_bootstap} as the 2D tracking model and TAPIP3D~\cite{tapip3d} as the 3D tracking model. The BootsTAP model outputs two logits for calculating confidence: occlusion logit $o$ and uncertainty logit $u$. We convert these raw logits into interpretable probability values as $v = 1 - \sigma(o)$ and $p = 1 - \sigma(u)$, where $v$ represents the visibility probability, $p$ denotes the certainty probability, and $\sigma(\cdot)$ is the sigmoid function.

To compute the final confidence $c$ for each track point, we multiply its visibility and certainty as $c = v \times p$. This calculated confidence represents how certain the 2D tracker is about the given track point. We filter the points by applying a threshold as shown in \cref{eq:mask}. We set the default threshold $\tau$ to 0.5. This value can be adjusted based on the scene complexity and the reliability of tracking results when adopting new 2D or 3D tracking models in the future, which would yield more stable results. Ablation study on threshold is in \cref{sec:add_ablation}.

As mentioned in \cref{subsec:initialization}, we employ $\mathcal{P}^{3D}$ with a transformer architecture, which simultaneously infers points within a window of fixed length $L$ frames. We set $L$ to 16 throughout our experiments. Each window undergoes 6 inference iterations. After each iteration, if there exist unprojected 2D track points with high confidence, they replace the corresponding points and serve as anchors. Based on these updated anchors, the 3D tracking model infers other points within the window in the next iteration step. We utilize dynamic object masks to exclude inferred points whose positions fall outside the mask region. We first run the 2D tracking model over all frames. The query points of 2D tracks at each timestep $t$ are then reused as query points for the 3D tracking model after unprojection. Using these query points together with the anchors, AnchorTAP3D generates 3D tracks for initialization. AnchorTAP3D takes 71 minutes for 130-frame video using an NVIDIA RTX A6000 D6 48GB GPU.

\subsection{Additional initialization details}
\label{subsec:additonal_initialization_details}
As mentioned in the initialization details in \cref{subsec:initialization}, we randomly select a subset of tracks across all query times from the 3D tracks obtained by AnchorTAP3D to initialize the Gaussians. Each track stores the position of a point at each frame, which we transform into the temporal trajectory of a single Gaussian in our model. Thus, one track corresponds to one Gaussian and its motion path over time. The initial color of each Gaussian is set by taking the point's color at the query time of the track.

\subsection{Optimization details}
\label{subsec:optimization_details}

We use the Adam~\cite{Kingma2014AdamAM} optimizer. In~\cref{eq:total_loss}, D-SSIM~\cite{article_ssim} loss and LPIPS~\cite{zhang2018unreasonable} loss are used together with $\mathcal{L}_{\text{rgb}}$ with weights of 0.2, 0.01, and 0.8, respectively. We apply $\mathcal{L}_{\text{mask}}$ to align rendered and predicted masks using MSE (weight 1.0). $\mathcal{L}_{\text{depth}}$ enforces depth consistency through an MSE term (weight 0.5) and smoothness via gradient regularization (weight 1.0). $\mathcal{L}_{\text{track}}$ ensures rendered trajectories match 2D track observations through two components: a 2D coordinate alignment term $\mathcal{L}_{\text{track-2d}}$ (weight 2.0) and a depth consistency term $\mathcal{L}_{\text{track-depth}}$ (weight 0.1) that matches reprojected and estimated depths. 

As mentioned in \cref{eq:arap}, $\mathcal{L}_{\text{arap}}$ is computed based on spatial relationships between temporally non-adjacent frames within each batch. In \cref{eq:arap}, $\mathcal{N}$ represents node pairs $(i,j)$, which are constructed in two ways. First, for each node $i$, we select the 5 nearest nodes across all frames. Second, to ensure that nodes aligned along the ray direction maintain consistent distances when representing the same rigid object, we select the 2 nodes with the most similar velocities along the ray direction and incorporate them into $\mathcal{N}$ to preserve their distances across different time frames. The nearest node method is applied throughout all training steps with a weight of 2.0.

We optimize Node-based motion representation and canonical Gaussians. For Gaussian parameters, we set learning rates as follows: mean (1.6 × 10$^{-4}$), opacity (1 × 10$^{-2}$), scale (5 × 10$^{-3}$), rotation (1 × 10$^{-3}$), and color (1 × 10$^{-2}$). We incorporate adaptive density control for Gaussians during training proposed in 3DGS~\cite{kerbl20233d}. Motion bases are optimized with a learning rate of 1.6 × 10$^{-4}$. For node parameters, we configure learning rates at 1.6 × 10$^{-5}$ for position, 5 × 10$^{-4}$ for radius, and 1 × 10$^{-2}$ for motion coefficients. We set the number of first-level nodes to $70$, second-level nodes to $10$, and the number of node bases to $15$ for iPhone360 experiments. Since there is no universally optimal value for these hyperparameters, as the ideal setting varies depending on scene characteristics, we report results under fixed values rather than per-scene tuning.

Since our model performs 360° reconstruction using a loss closely related to prior works~\cite{liang2025himor,wang2025shape}, the overall optimization pipeline does not introduce substantial additional computational overhead compared to existing approaches. Our model requires approximately 3 hours to reconstruct on 140 images using an NVIDIA RTX A6000 D6 48GB GPU.

\begin{figure}[t]
    \centering
    \includegraphics[width=\linewidth]{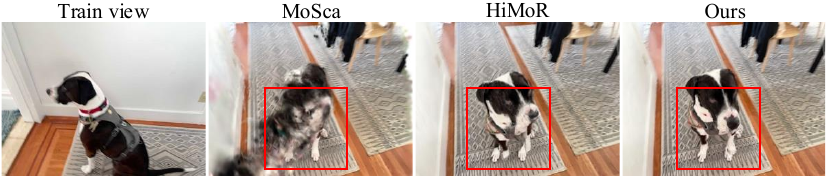}
    \caption{
    iPhone dataset~\cite{gao2022dycheck} \textit{Haru-sit} scene(no test ground truth) qualitative result. It shows that our model reconstructs unseen regions such as the leg area with reasonable accuracy.
    }
    \label{fig:harusit}
\end{figure}

\section{Additional Results}
\label{sec:add_results}
\subsection{Evaluation details}
\label{subsec:evaluation_details}
\begin{figure}[t]
    \centering
    \includegraphics[width=\linewidth]{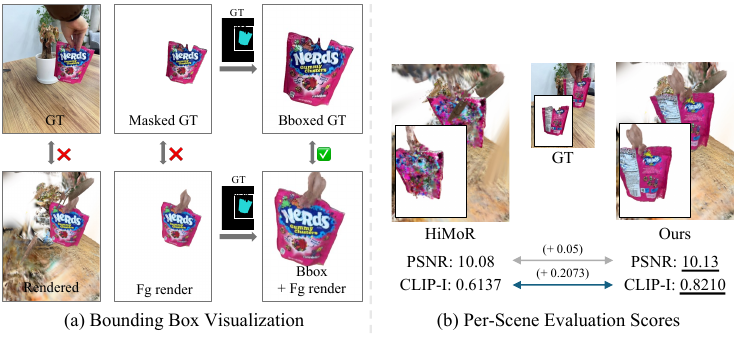}
    \caption{
    (a) shows Visualization of bounding box (Bbox) application. For both the GT and rendered images, the bounding boxes are computed from a processed version of the GT mask. Evaluation between GT and rendered images is unreliable due to background ambiguity. Evaluation between masked GT and foreground rendering is also problematic due to excessive white regions. Evaluation between Bbox + GT and Bbox + FG rendering provides the most reliable comparison strategy. (b) denotes per-scene evaluation scores to illustrate the correspondence between quantitative metrics and visual results, where CLIP-I better reflects the perceptual similarity compared to the pixel-wise metric PSNR.
    }
    \label{fig:suppl_bbox_score}
\end{figure}

As mentioned in~\cref{sec:experiments}, we adopt bounding boxes to compare with ground truth at extreme novel-view synthesis. When synthesizing extreme novel views in object-centric scenes, background regions appear mostly white because training cameras rarely observe those areas. Therefore, to exclude unseen background regions from evaluation and focus on the reconstruction of dynamic objects, we compare renders using only the dynamic foreground Gaussians against the corresponding dynamic object regions in the test images. However, naively computing CLIP scores between foreground Gaussian renders and foreground-masked ground truth is ineffective due to excessive white regions in both images. Instead, we create bounding boxes by extending the ground truth mask horizontally and vertically by 1/8 of the image resolution on each side. We apply these bounding boxes to both images and compute CLIP similarity scores within the cropped regions. Visualizations before and after bounding box application are shown in~\cref{fig:suppl_bbox_score}(a).

We also provide ground truth and render image pairs and per-scene quantitative scores in~\cref{fig:suppl_bbox_score}(b), illustrating the correlation between the magnitude of scores and the visual results. As can be observed, the actual extreme novel-view synthesis results of HiMoR~\cite{liang2025himor} and Ours exhibit substantial perceptual differences, which is reflected in the CLIP-I scores. This discrepancy is not clearly captured by pixel-wise evaluation metrics. Furthermore, the absolute values of pixel-wise scores are relatively lower compared to those in conventional tasks, as the evaluation is performed on extreme novel view. As mentioned in~\cref{sec:experiments}, monocular dynamic scene reconstruction under novel-view synthesis is a highly ill-posed problem. In particular, when it comes to extreme novel-view, most object regions visible from extreme viewpoints are not observed in the training images at the same timestamp, making it difficult to precisely guide object localization. Combined with the sensitivity of pixel-wise metrics to even small positional errors, achieving high scores becomes inherently challenging, and the scores may not always align with human perceptual quality. Pixel-wise evaluations are reported in~\cref{tab:pixel_eval} for models supporting 1x resolution inference, and all training and evaluation on the iPhone and iPhone360 datasets are conducted at 1x resolution.

\subsection{iPhone360 dataset}
\label{subsec:iphone360_dataset}
We provide a \textbf{Supplementary Video} at project page for qualitative validation. It includes bullet-time effect 360° results and extreme test view comparisons with ground truth. The bullet-time effect 360° results demonstrate that our model can reconstruct 360° of an object at every dynamic time step if it appears in the video. These results demonstrate that our model surpasses baseline models in 360° reconstruction~\cite{wang2025shape, liang2025himor, lei2025mosca}.
In the \textit{jelly}, \textit{jacket}, and \textit{goat} scenes, only our method produces proper geometry. The baselines exhibit abnormal geometric deformations, simply overfitting to the training video.
In the \textit{walk-around} scene, the baselines~\cite{lei2025mosca, liang2025himor} exhibit clear visual artifacts and abrupt velocity changes. Gaussians originating in regions visible to the training cameras undergo abrupt velocity changes when they pass through occluded areas. Consequently, the person's lower body appears disconnected and exhibits unnatural motion with abrupt velocity changes. These results indicate that when initialization is based on 2D tracks, optimization fails to reliably reconstruct invisible regions and instead induces sharp velocity changes to overfit the training images.
As also shown in the bullet-time 360° results on the \textit{goat} scene, the baseline~\cite{liang2025himor} struggles to reconstruct even the training views under complex dynamic motion.
In contrast, our method places Gaussians in plausible locations within invisible regions from the outset via AnchorTAP3D, effectively preventing Gaussian drifting even under a similar optimization scheme.

\begin{figure}[t]
    \centering
    \includegraphics[width=\linewidth]{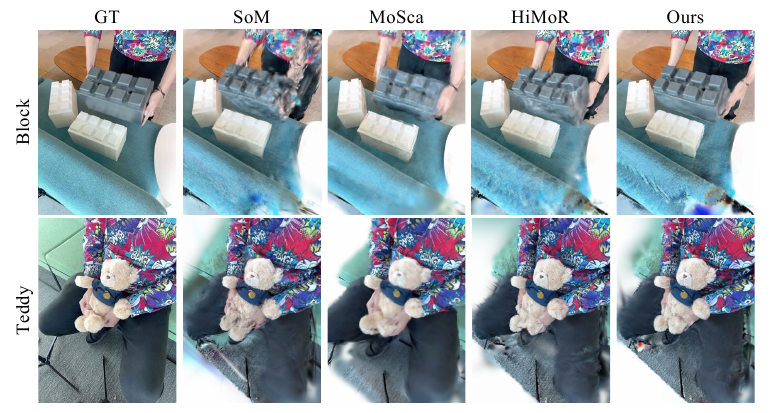}
    \vspace{-1em}
    \caption{
    Qualitative comparison with the baseline models~\cite{wang2025shape, lei2025mosca, liang2025himor} on the iPhone dataset~\cite{gao2022dycheck} Block and Teddy scenes.
    }
    \label{fig:suppl_iphone}
\end{figure}

\subsection{iPhone dataset}
\label{subsec:iphone_dataset}
The \textit{Haru-sit} scene in the iPhone dataset~\cite{gao2022dycheck} captures various parts of the object through a moving camera. Since \textit{Haru-sit} scene lacks novel view ground truth, we report qualitative performance comparisons with existing models on a new pseudo test view in \cref{fig:harusit}, demonstrating our model's capability in reconstructing regions far from the training view. 
Comparisons on the \textit{Block} and \textit{Teddy} scenes are shown in~\cref{fig:suppl_iphone}. As the iPhone dataset involves relatively smaller view discrepancy compared to iPhone360, baseline models also produce competitive results in these scenes.

\begin{figure}[t]
    \centering
    \includegraphics[width=\linewidth]{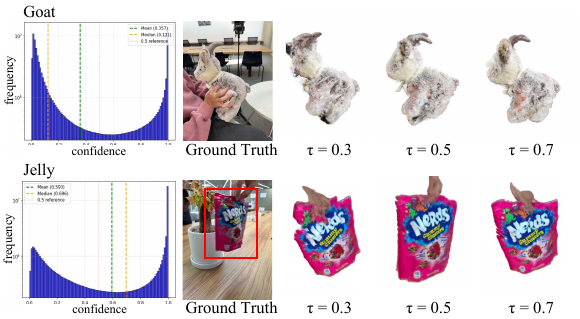}
    \caption{
    \textbf{AnchorTAP3D Threshold Ablation.} The left graph shows the confidence distribution of 2D tracking points for each scene, with the mean and median confidence values displayed at the top. The right side presents the final 4DGS360 rendering results with varying $\tau$ values applied during the AnchorTAP3D process. This allows us to observe the correlation between the confidence distribution of 2D tracking results and the choice of threshold.
    }
    \label{fig:t_ablation}
\end{figure}

\begin{figure}[t]
    \centering
    \setlength{\tabcolsep}{4pt}
        \begin{minipage}[t]{0.49\columnwidth}
        \centering
        \captionof{table}{Quantitative results on AnchorTAP3D ablation.}
        \label{tab:quant_anchor_ablation}
        \resizebox{\linewidth}{!}{%
        \begin{tabular}{l cccc}
            \toprule
            Method & PSNR$\uparrow$ & SSIM$\uparrow$ & LPIPS$\downarrow$ & CLIP-I$\uparrow$ \\
            \midrule
            w/o 3D init       & 10.990 & 0.685 & 0.319 & 0.798 \\
            w/o Anchor   & 11.282 & 0.695 & 0.402 & 0.875 \\
            Ours              & \textbf{11.336} & \textbf{0.701} & \textbf{0.287} & \textbf{0.882} \\
            \bottomrule
        \end{tabular}}
    \end{minipage}
    \hfill
    \begin{minipage}[t]{0.49\columnwidth}
        \centering
        \captionof{table}{Quantitative results (CLIP-I scores) on the DIFIX3D+ ablation.}
        \label{tab:quant_difix_ablation}
        \begin{tabular}{l cc}
            \toprule
            Method & Jacket & Jelly \\
            \midrule
            HiMoR              & 0.70 & 0.68 \\
            HiMoR w/ DIFIX3D+  & 0.84 & 0.72 \\
            Ours               & \underline{0.92} & \underline{0.83} \\
            Ours w/ DIFIX3D+   & \textbf{0.94} & \textbf{0.85} \\
            \bottomrule
        \end{tabular}
    \end{minipage}
\end{figure}

\section{Additional Ablation}
\label{sec:add_ablation}

\subsection{AnchorTAP3D threshold ablation}
\label{subsec:anchor_threshold_ablation}
As noted in~\cref{subsec:anchortap3d_details}, all experiments are conducted with a fixed threshold of $\tau = 0.5$ to evaluate the general performance of our model. Here, we provide threshold ablations in~\cref{fig:t_ablation} to further investigate how the choice of threshold interacts with the confidence distribution of 2D tracking results across different scenes. The \textit{goat} scene has a mean confidence of $0.357$, while the \textit{jelly} scene has a mean of $0.593$, indicating that the 2D tracker finds the \textit{goat} scene more challenging and the \textit{jelly} scene more tractable. More reliable 2D tracking results suggest that setting a lower threshold to incorporate more 2D tracking results as anchors can still be beneficial in certain scenes. Consequently, as reflected in the final rendering results, the \textit{jelly} scene achieves its best performance at $\tau = 0.5$, while the \textit{goat} scene benefits from a higher threshold of $\tau = 0.7$. Note that excessively low thresholds such as $\tau = 0.3$ also lead to degraded results for both scenes. For practitioners seeking to reconstruct new scenes, we suggest setting $\tau \approx 1 - \text{ratio}(\text{confidence} > 0.5)$ as a reasonable initial threshold to begin the search for an optimal value.

\subsection{In-the-wild depth estimation ablation}
\begin{figure}[t]
    \centering
    \includegraphics[width=\linewidth]{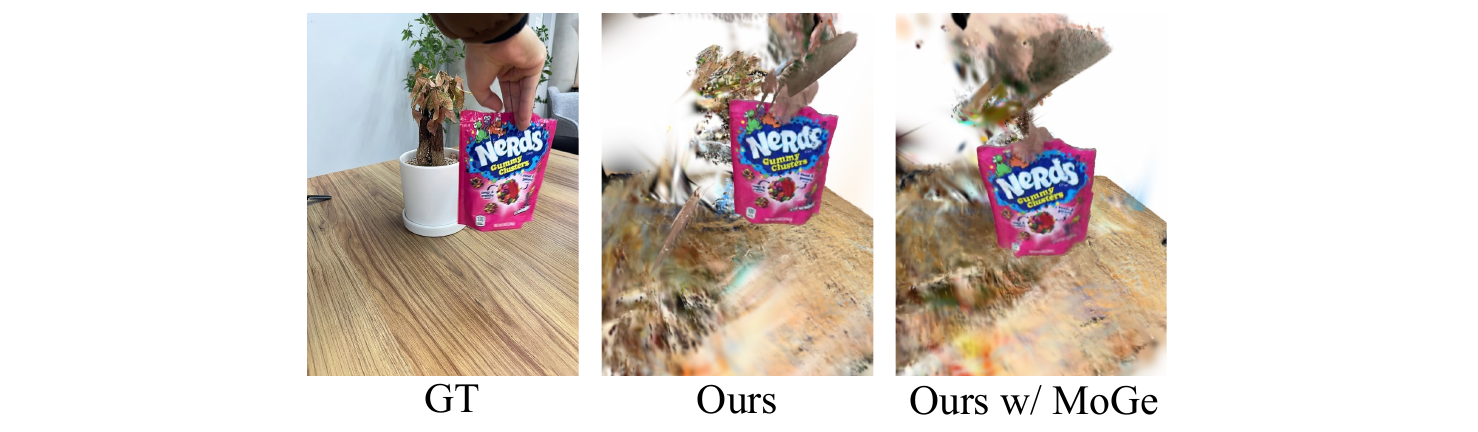}
    \caption{
    \textbf{Estimated depth results}. 'Ours' denotes the final reconstruction result using depth estimated by a depth estimation model and corrected with LiDAR sensor depth, while 'Ours w/ MoGe' denotes the result using only the depth estimation model MoGe~\cite{wang2026moge}.
    }
    \label{fig:suppl_estdepth}
\end{figure}

As mentioned in \cref{sec:experiments}, results on the iPhone360 dataset use LiDAR sensor values to correct the estimated depth. \cref{fig:suppl_estdepth} reports the reconstruction results on the jelly scene from the iPhone360 dataset using only the depth inferred by the depth estimation model, without LiDAR correction. These results show that while the object reconstruction quality itself is preserved in `w/ MoGe', the MoGe-only reconstruction lies at an arbitrary scale and is not spatially aligned with the ground truth at the test camera.

\subsection{More quantitative ablations}
In \cref{tab:quant_anchor_ablation}, we quantitatively report the experimental results from \cref{fig:ablation}, conducted on the iPhone360 dataset. Additionally, in \cref{tab:quant_difix_ablation}, similar to what was done in \cref{fig:difix}, we quantitatively report the ablation results of applying DIFIX3D+ to HiMoR and Ours, respectively, on the \textit{Jacket} and \textit{Jelly} scenes from the iPhone360 dataset.

\section{iPhone360 Dataset Overview}
\label{sec:iphone360_overview}
The supplementary video provides an overview of the scenes comprising the iPhone360 dataset. The \textit{goat}, \textit{jelly}, and \textit{walk-around} scenes are captured with one training camera and two test cameras, while \textit{block2}, \textit{jacket}, and \textit{pull-up} are captured with one training camera and one test camera. In the \textit{goat}, \textit{jacket}, \textit{jelly}, and \textit{walk-around} scenes, the training and test cameras consistently maintain an angular difference of over 90°. For \textit{block2} and \textit{pull-up}, the training camera moves more actively, resulting in a varying angular difference with respect to the test camera. The six scenes cover a diverse range of motions, including rigid object movement, deformable objects, and human motion. Furthermore, all scenes are captured with a hand-held moving camera, faithfully reflecting real-world capture scenarios. The iPhone360 dataset is the first dataset designed to evaluate monocular dynamic reconstruction under extreme novel-view synthesis in real-world capture scenarios.


\clearpage

\end{document}